\documentclass[10pt,twocolumn,letterpaper]{article}

\usepackage[pagenumbers]{cvpr} %

\usepackage[accsupp]{axessibility} %

\makeatletter
\DeclareRobustCommand\onedot{\futurelet\@let@token\@onedot}
\def\@onedot{\ifx\@let@token.\else.\null\fi\xspace}
\def\eg{\emph{e.g}\onedot}

\def\wrt{w.r.t\onedot} 
\def\etal{\emph{et al}\onedot}
\makeatother

\usepackage{microtype}
\usepackage{epsfig}
\usepackage{graphicx}
\usepackage{amsmath}
\usepackage{amssymb}
\usepackage{xcolor}
\usepackage{caption}
\usepackage{booktabs}

\usepackage{wrapfig}
\usepackage{multirow}

\usepackage{textcomp}

\usepackage{gensymb}
\usepackage{xspace}
\usepackage[ruled,vlined]{algorithm2e}
\usepackage[super]{nth}
\usepackage{cuted} %

\definecolor{turquoise}{cmyk}{0.65,0,0.1,0.3}
\definecolor{purple}{rgb}{0.65,0,0.65}
\definecolor{dark_green}{rgb}{0, 0.5, 0}
\definecolor{orange}{rgb}{0.8, 0.6, 0.2}
\definecolor{red}{rgb}{0.8, 0.2, 0.2}
\definecolor{blueish}{rgb}{0.0, 0.7, 1}
\definecolor{light_gray}{rgb}{0.7, 0.7, .7}
\definecolor{pink}{rgb}{1, 0, 1}
\definecolor{dark_red}{rgb}{0.5, 0, 0}
\definecolor{pptblue}{RGB}{68, 114, 196}
\definecolor{pptyellow}{RGB}{255, 192, 0}
\definecolor{pptblue}{RGB}{68, 114, 196}
\definecolor{pptgreen}{RGB}{112, 173, 71}
\definecolor{pptorange}{RGB}{237, 125, 49}
\definecolor{pptred}{RGB}{237, 70, 49}

\definecolor{r10}{rgb}{0.400, 0.761, 0.647}
\definecolor{r20}{rgb}{0.988, 0.553, 0.384}
\definecolor{r30}{rgb}{0.553, 0.627, 0.796}

\usepackage{pifont}
\usepackage{xcolor}

\def\real{\mathbb{R}}

\usepackage{bbm}
\def\indicator{\mathbbm{1}}

\newcommand{\revised}[1]{{#1}}
\newcommand{\approachName}{DS}

\newcommand{\Paragraph}[1]{\vspace{1.5mm} \noindent \textbf{#1} \hspace{0mm}}

\usepackage[pagebackref,breaklinks,colorlinks]{hyperref}
\hypersetup{
    colorlinks=true,
    linkcolor=pptred,
    filecolor=pptblue,      
    urlcolor=pptorange,
    citecolor=pptblue,
}

\usepackage[capitalize]{cleveref}
\crefname{section}{Sec.}{Secs.}
\Crefname{section}{Section}{Sections}
\Crefname{table}{Table}{Tables}
\crefname{table}{Tab.}{Tabs.}

\def\cvprPaperID{6252} %
\def\confName{CVPR}
\def\confYear{2022}

\begin{document}

\title{Differentiable Stereopsis: Meshes from multiple views \\ using differentiable rendering}

\author{Shubham Goel\\
UC Berkeley\\
{\tt\small shubham-goel@berkeley.edu}
\and
Georgia Gkioxari\\
Meta AI\\
{\tt\small gkioxari@fb.com}
\and
Jitendra Malik\\
UC Berkeley\\
{\tt\small malik@eecs.berkeley.edu}
}
\maketitle

\begin{abstract}
   We propose Differentiable Stereopsis, a multi-view stereo approach that reconstructs shape and texture from few input views and noisy cameras. We pair traditional stereopsis and modern differentiable rendering to build an end-to-end model which predicts textured 3D meshes of objects with varying topologies and shape. We frame stereopsis as an optimization problem and simultaneously update shape and cameras via simple gradient descent. We run an extensive quantitative analysis and compare to traditional multi-view stereo techniques and state-of-the-art learning based methods. We show compelling reconstructions on challenging real-world scenes and for an abundance of object types with complex shape, topology and texture. \footnote{Project webpage: \url{https://shubham-goel.github.io/ds/}}
\end{abstract}

\begin{figure}
\begin{center}
   \includegraphics[width=0.9\linewidth]{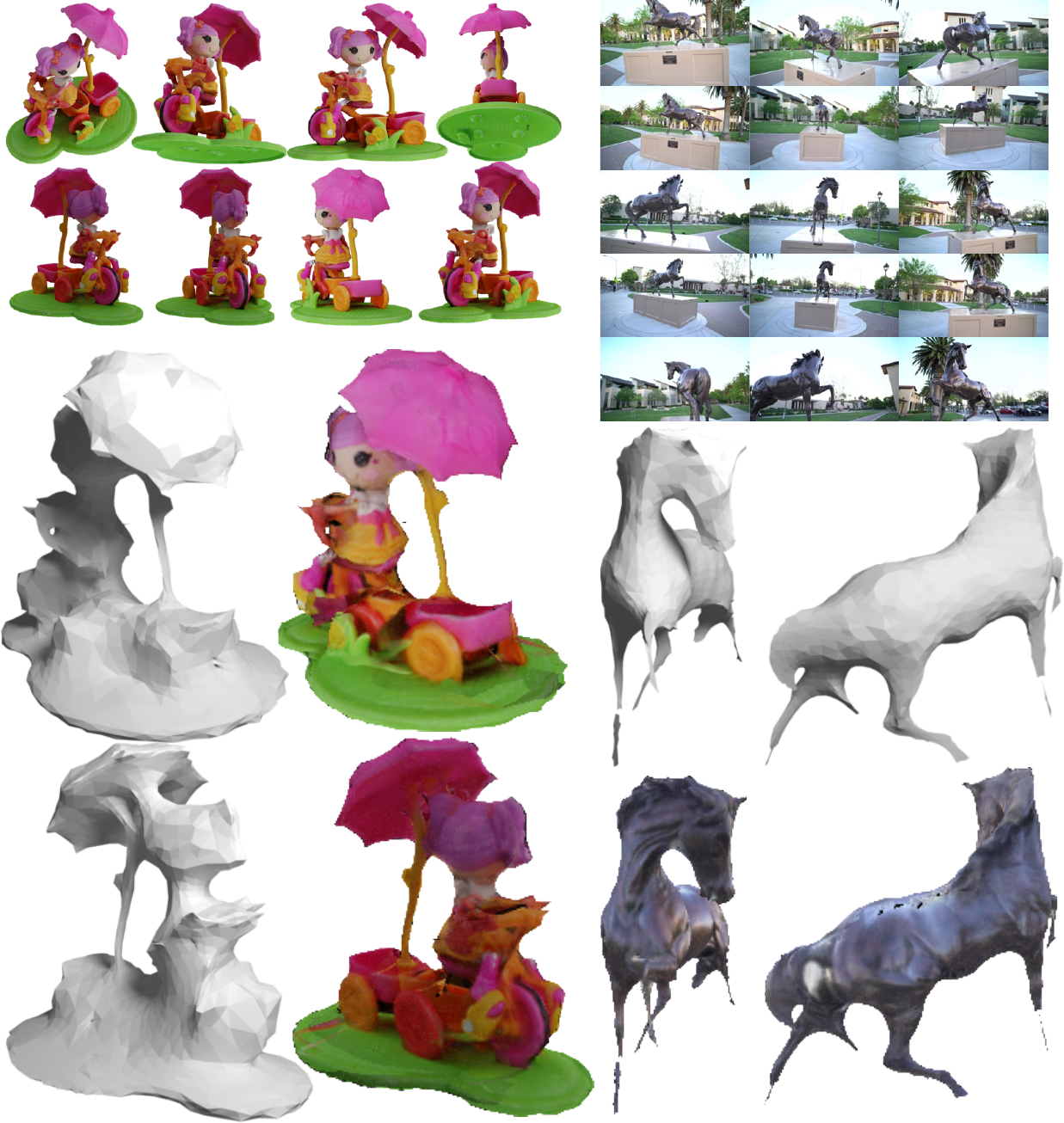}
\end{center}
   \vspace{-5mm} %
   \caption{Reconstructions with Differentiable Stereopsis (\approachName) from few input views and noisy cameras. We show input views (top) and novel views of reconstructions (bottom).}
   \vspace{-5mm} %
\label{fig:teaser}
\end{figure}

\vspace{-2mm}
\section{Introduction}

Binocular stereopsis~\cite{wheatstone1838}, and its multi-view cousin, Structure from Motion~\cite{hartley2003multiple, szeliski2010computer}, has traditionally been formulated as a two stage process:
\begin{enumerate}
\item Find corresponding 2D points across views, which are the 2D projections of the \emph{same} 3D scene point. 
\item Recover relative orientations of cameras, and the depths of these points by triangulation.
\end{enumerate}
In this work, we bypass the first stage of finding point correspondences across images and directly estimate 3D shape and cameras given multiple 2D views \revised{with noisy cameras}. 
We formulate this as an optimization problem that we solve using newly developed differentiable rendering tools. We name our approach \emph{Differentiable Stereopsis}.

Our approach is linked to old work in multi-view geometry, and in particular model-based stereopsis which was explored by Debevec~\etal~\cite{debevec1996modeling} and related ideas in plane plus parallax by Irani~\etal~\cite{irani2002direct}.
The key observation in model-based stereo is simple: two images of the same scene which appear different become similar after projection onto an approximate 3D model of the scene.
Projecting the texture from one image onto the 3D model produces a warped version of that view which when transformed from a second view is directly comparable to the second image.
Initially, the 3D model and the estimated relative camera orientation are inaccurate.
But as shape and camera predictions improve, the two images will start to look more similar and will eventually become identical -- in the idealized case of Lambertian surfaces and no imaging noise.
Upon convergence, the shape is expected to be an accurate representation of the scene.

\begin{algorithm}[t!]
\SetAlgoLined
 \textbf{Input:}  $I_{1,2}$, $\pi_{1,2}$\;
 $S \gets $ Sphere()\;
 \While{ not converged}{
    points $\gets$ \texttt{rasterize}($S$, $\pi_1$)\;
    texels $\gets$ \texttt{sample}($I_2$, $\pi_2$(points))\;
    $I^r_1$ $\gets$ \texttt{blend}(texels)\;
    loss $\gets$ \texttt{compute\_loss}($I^r_1$, $I_1$)\;
    $S,\pi_{1,2}$ $\gets$ $S,\pi_{1,2} -$ lr * \texttt{gradient}(loss)\;
 }
 \textbf{Outputs} $S$, $\pi_{1,2}$
 \caption{Differentiable Stereopsis (2-view)}
 \label{algo:approach}
\end{algorithm}

An important step in traditional stereopsis is finding 2D correspondences across views.
We bypass this and directly recover shape and cameras using modern optimization techniques.
We frame stereopsis as an optimization problem by minimizing a differentiable objective with respect to shape and cameras.
To this end, we exploit advances in \emph{differentiable rendering}~\cite{loper2014opendr, NMR, liu2019soft, chen2019learning, ravi2020pytorch3d} to project shape and texture onto image planes which we compare to scene views.\linebreak
We update shape and cameras via gradient descent.
Algorithm~\ref{algo:approach} illustrates our proposed Differentiable Stereopsis (\approachName) for the case of 2 views.
\revised{We rely on (i) object masks to isolate and refine object topology; and (ii) noisy camera pose initializations, which may still come from a correspondence matching algorithm.}
Fig.~\ref{fig:teaser} shows shape reconstructions with \approachName~when noisily posed views are given as input.

At the core of our approach is a novel and differentiable texture transfer method which pairs rendering with the key insight of texture warping via 3D unprojections.
Our texture transfer learns to sample texture from the input views based on the shape estimate and noisy cameras. 
To allow for differentiation, it composites the final texture in a soft manner by weighing texture samples proportionally to their visibility and direction from each view.

We test our approach on challenging datasets and on a large variety of objects with complex and varying shapes.
Unlike prior works that assume several dozens of object views, our experimental settings follow real-world practical scenarios where only a few views are available. 
For example, Amazon, eBay or Facebook Marketplace only contain a handful of views for each listed item, and any 3D reconstruction has to originate from 10 views or less.
We emphasize on this harder, yet more realistic, setting and show empirical results with real product images from Amazon~\cite{collins2022abo}.
On Google's Scanned Objects~\cite{google2020scanned} we perform an extensive quantitative and qualitative analysis and compare to competing approaches %
under settings similar to ours. 
We also show results on Tanks and Temples~\cite{knapitsch2017tankstemples} which contains RGB views of complex scenes, as shown on the right in Fig.~\ref{fig:teaser}.

\section{Related Work}

Extracting 3D structure from 2D views of a scene is a long standing goal of computer vision. 
Classical multi-view stereo methods and Structure from Motion techniques~\cite{hartley2003multiple, szeliski2010computer, debevec1996modeling, furukawa2009accurate, irani2002direct} find correspondences across images and triangulate them into points in 3D space. 
The resulting point clouds, if dense enough, can be meshed into surfaces \cite{bernardini1999ball, kazhdan2006poisson}. 
The culmination of a long line of classical SfM and stereo approaches is COLMAP~\cite{schoenberger2016sfm, schoenberger2016mvs} -- a widely used tool for estimating camera poses and reconstructing dense point clouds from 2D input views.
All aforementioned techniques assume calibrated and accurate cameras and thus are not very robust to camera noise.

Finding point correspondences, the first stage of stereopsis, is challenging especially in the case of sparse widely-separated views.
Debevec~\etal~\cite{debevec1996modeling} tackle this by proposing model-based stereopsis wherein a coarse scene geometry allows views to be placed in a common reference frame, making the correspondence problem easier. 
We draw inspiration from this work and pair it with new learning tools to reconstruct textured 3D meshes from sparse views.
We frame stereopsis as an optimization problem and minimize a differentiable objective which allows both shape and cameras to self-correct.
This increases robustness to camera noise, in antithesis to classical techniques. 

Recent work on multi-view stereo~\cite{yao2018mvsnet, yan2020dense} train deep neural nets (DNNs) with depth supervision.
As expected, these methods outperform COLMAP for point cloud reconstruction but are limited as they need ground truth.
We rely solely on image re-projection losses and no true depth information.

Work on unsupervised depth prediction \cite{zhou2017unsupervised, li2018megadepth, yin2018geonet,zhang2020unsupervised} estimate depth via DNNs trained on monocular videos and without ground truth depth.
They exploit photometric and depth consistency across multiple views, much like classical stereo. 
However, they focus on forward-facing scenes like KITTI~\cite{geiger2013vision} and do not reconstruct high-fidelity shape.

There is extensive work on recovering shape from images using differentiable rendering~\cite{NMR, loper2014opendr, liu2019soft, chen2019learning, ravi2020pytorch3d, kanazawa2018learning, goel2020shape, li2020self, tulsiani2020implicit, kulkarni2020articulation, ye2021shelf}. These approaches focus on extracting object priors by training on large datasets and test on images of seen categories.
We also use differentiable rendering~\cite{NMR, loper2014opendr, liu2019soft, chen2019learning, ravi2020pytorch3d} to frame stereopsis as a differentiable optimization problem.
Differentiating with respect to shape and camera allows for both to self-correct during optimization.

Most relevant to our work are methods that learn shape by fitting to a set of images. Early work on extracting shape from silhouettes used a visual hull \cite{laurentini1994visual}. Gadelha~\etal~\cite{gadelha20173d} reconstruct voxels from silhouettes and noisy camera poses via differentiable projection but don't use any texture information. 
However, shape details such as concavities cannot be captured by silhouettes.
We show in Fig.~\ref{fig:qual-notex} in our experiments (Sec.~\ref{sec:exp}) that optimizing for shape without texture information fails to reconstruct creases in shape. 
{Some variational approaches for MVS \cite{faugeras1998complete, pons2007multi, hiep2009towards, delaunoy2014photometric} exploit photometric consistency to refine shape via gradient descent but they require many images, initial shapes or accurate cameras}. 
Recently, IDR~\cite{yariv2020multiview} and DVR~\cite{niemeyer2020differentiable} recover shape from multiple posed images and masks using implicit volumetric representations. IDR shows superior results to DVR and claims to work with few input views and slightly noisy camera poses; a setting similar to ours. We compare to IDR in Sec.~\ref{sec:exp}.

Recent novel-view-synthesis approaches \cite{zhou2018stereo, lombardi2019neural, mildenhall2020nerf} encode volumetric occupancy information in their internal representations for the task of image synthesis from novel viewpoints.
While they don't explicitly learn shape, their representation can be processed to extract geometry. 
NeRF~\cite{mildenhall2020nerf} is one such approach which takes posed multiple views as input and encodes occupancy and color for points in 3D space as an implicit function. 
In Sec.~\ref{sec:exp}, we compare to NeRF and extend it with a variant that optimizes for noisy cameras by enabling backpropagation to its parameters.

\section{Approach}
\newcommand{\visprob}{\sigma}
\newcommand{\cosprob}{\gamma}
\newcommand{\hypervisprob}{\tau_\text{vis}}
\newcommand{\hypercosprob}{\tau_\text{cos}}

\newcommand{\loss}[1]{L_\text{#1}}

\begin{figure}[t]
\begin{center}
   \includegraphics[width=0.99\linewidth]{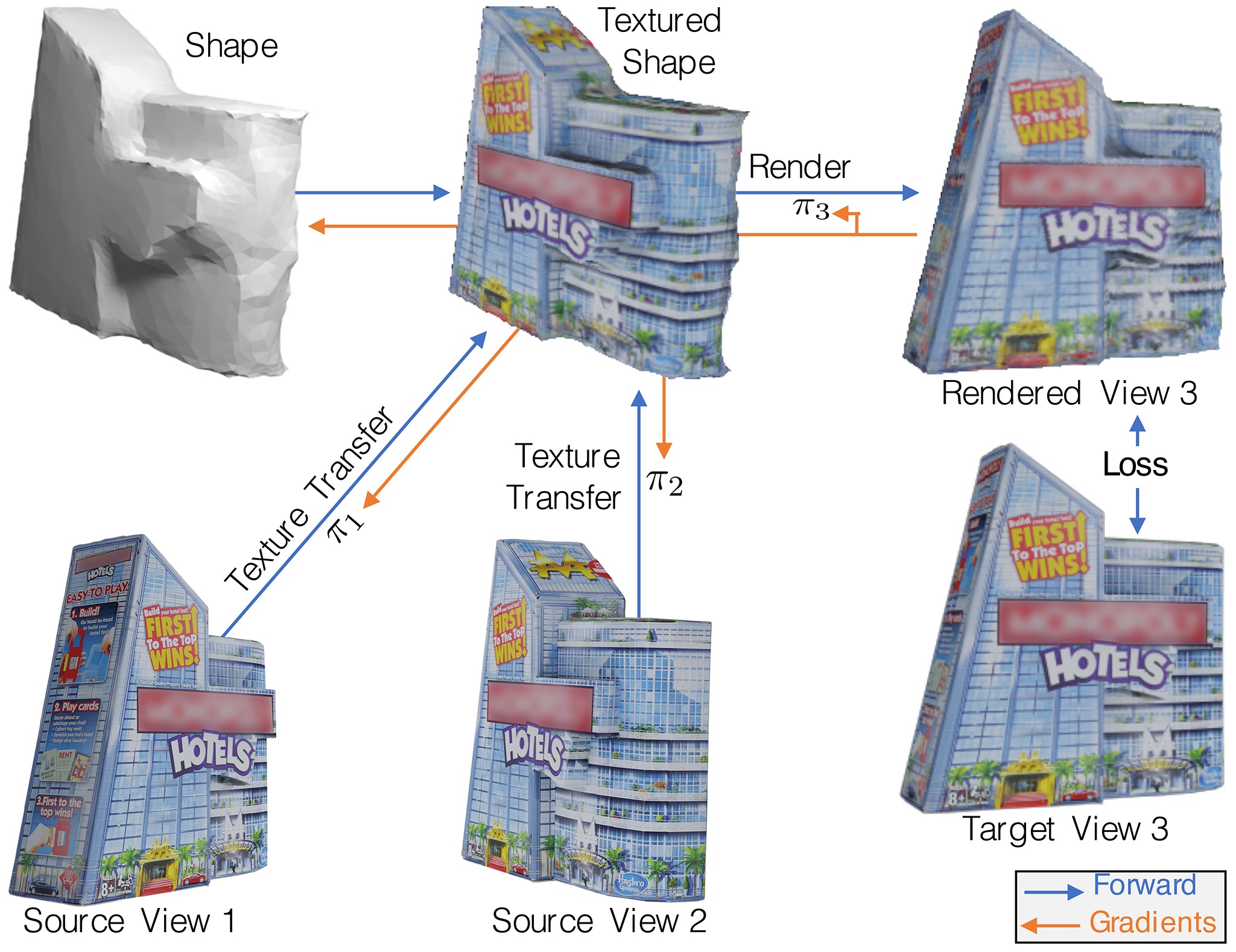}
\end{center}
   \vspace{-4mm} %
   \caption{Overview of Differentiable Stereopsis (\approachName).
   The shape estimate is textured using source views, rendered from a target view's camera and compared against the target view. The loss is backpropagated to update shape and cameras.
   }
   \label{fig:approach}
   \vspace{-4mm} %
\end{figure}

We tackle the problem of stereopsis using modern differentiable rendering techniques.
Our approach takes $N$ image views $I_{1..N}$ of an object with corresponding masks $A_{1..N}$ and \textit{noisy} camera poses $\pi_{1..N}$ as input, and outputs the shape of the object as a textured mesh. 
We frame stereopsis as an optimization problem, outlined in Fig.~\ref{fig:approach}.
We iteratively render a shape estimate from multiple cameras using differentiable textured rendering and update the shape and cameras by minimizing image reprojection losses. 

We first provide some background on differentiable rendering and then describe our approach in detail.
\subsection{Background}
We define a textured mesh $M = (V, F, T)$ as a set of vertices $V$, faces $F$ and a texture map $T$.
Under a camera viewpoint $\pi$, mesh $M$ is rendered to image $I^r= R_T(M, \pi)$ and mask $A^r=R_S(M, \pi)$, where $R_T$ denotes textured rendering and $R_S$ silhouette rendering.

Both $R_S$ and $R_T$ perform mesh rasterization.
Rasterization computes which parts of the mesh are projected to a pixel on the image plane. 
For each pixel $p$, we find the $K$ nearest faces that intersect with a ray originated at $p$~\cite{ravi2020pytorch3d}.

In the case of silhouette rendering $R_S$, rasterization is followed by a soft silhouette shader. 
This shader assigns each pixel an occupancy probability by blending the euclidean distance of the pixel to each of the $K$ faces~\cite{liu2019soft, ravi2020pytorch3d}. 

For textured rendering $R_T$, we use a texture shader which computes the RGB color for each pixel $p$ in the image. 
This shader blends the colors from the top $K$ faces for each pixel, as computed by the rasterizer.
For the $k$-th face, the color $c_k = T(x)$ is computed by sampling the texture map $T$ at the point of intersection $x$ of the ray originating at $p$ and the $k$-th face.
The set of colors $c_{1..K}$, also known as \textit{texels}, are composited to get the final color for the pixel.

\subsection{Texture Transfer}
\label{sec:tex-transfer}

The goal of our approach is to find $M=(V,F,T)$ that represents the object as seen from the noisily posed input views.
For each shape hypothesis $(V,F)$ we need to find the optimal texture $T$.
We introduce a \emph{novel} texture shader which relies on texture transfer from the inputs $I_{1...N}$.

Our shader computes the texture map $T$ as a function of the shape hypothesis $(V, F)$ and the posed input views $I_{1...N}$. 
The texture map $T:x \rightarrow (r,g, b)$ assigns an RGB color for each point $x$ on the mesh surface. 
The color is directly sampled from one or more input views. 
We build on a key insight: for a correct shape $(V,F)$ and correct cameras $\pi_{1...N}$, there exists one (or many) view $i$ in which $x$ is unoccluded, or in other words, there is a clear line-of-sight to $x$. 
For all such views, the projections $\pi_i(x)$ in the images correspond to the same 3D point $x$ and for Lambertian surfaces, all these points will share the same color $I_i(\pi_i(x))$.
The color $T(x)$ assigned to point $x$ is composited from the colors $I_i(\pi_i(x))$ for all views with a clear line-of-sight to $x$.
Formally, we define the texture transfer as follows:
\begin{equation}
T(x) = \sum_i w_i I_i(\pi_i(x))
\label{eq:textrans}
\end{equation}
where weights are unit-normalized and defined as $w_i = \visprob_i \cosprob_i$.

$\visprob$ encodes whether $x$ has a clear line-of-sight from the corresponding view.
Formally, we compare the z-distance of the camera transformed point $\pi_i(x)$ to the rendered depth map $D_i$ at $\pi_i(x)$ as follows
\begin{equation}
    \visprob_i = \exp ({-(\pi_i(x)_z - D_i(\pi_i(x)))/\hypervisprob})
\end{equation}
If there is a clear line-of-sight to $x$, then $\pi_i(x)_z \approx D_i(\pi_i(x))$ and thus $\visprob_i \approx 1.0$. 
If $x$ is obstructed by other parts of the shape, then $\pi_i(x)_z > D_i(\pi_i(x))$ and $\visprob_i < 1.0$.
We set the temperature $\hypervisprob$ to $10^{-4}$.

$\cosprob$ is a heuristic that favours views that look at $x$ fronto-parallel with minimal foreshortening. 
If $\hat{n}_i(x)$ is the outward surface-normal at $x$ in $i$-th view coordinates, then
\begin{equation}
    \cosprob_i = \indicator[\hat{n}_i(x)_z < 0] \exp( -(1 + \hat{n}_i(x)_z) / \hypercosprob )
\end{equation}
$\cosprob_i$ is highest when the normal points opposite the camera's z-axis, or $\hat{n}_i(x)_z = -1$. 
$\cosprob_i$ decreases exponentially as $\hat{n}_i(x)_z$ increases. 
We set the temperature $\hypercosprob$ to $0.1$.
In addition, we cull backward-facing normals ($\hat{n}_i(x)_z > 0$) to correctly sample texture in thin surfaces where $\visprob$ fails to capture visibility information of points on the two sides of the surface.

\Paragraph{Texture Rendering}
We described how to sample texture for a point $x$ on the mesh surface.
To render the texture under a viewpoint $\pi$, for each pixel $p$ we sample texels $c_{1..K}$ for all points $x_{1..K}$ with $c_k = T(x_k)$, where $x_k$ is the point on the $k$-th face that intersects the ray originating at $p$.
We use softmax blending~\cite{liu2019soft} to composite the final color at $p$.

\subsection{Optimization}
We have explained how to define the texture map $T$ for an object shape $(V, F)$ given posed input views $I_{1..N}$ and we have described how to render $M=(V, F, T)$ to images and silhouettes. 
We now describe our objective and how we optimize it \wrt vertices $V$ and cameras $\pi_{1..N}$.

\Paragraph{Parametrization} 
We parametrize a camera $\pi=(r, t, f)$ as rotation via an axis-angle representation $r \in \real^3$ (magnitude $|\mathbf{r}|$ is angle, normalization $\mathbf{r}/|\mathbf{r}|$ is axis), translation as $t \in \real^3$ and focal length $f$ as half field-of-view. 

We parametrize geometry as $V = V_0 + \Delta V$ where $\Delta V \in \real^{|V|\times3}$ is the deformation being optimized and $V_0$ are initial mesh vertices which remain constant.

\Paragraph{Objective}
Given a shape hypothesis $M=(V, F, T)$, cameras $\pi_{1..N}$ and input views $I_{1..N}$, we render silhouette $A^r_i=R_S(M, \pi_i)$ and image $I^r_i=R_T(M, \pi_i)$ for each view $i=1,...,N$.
We define our total loss to be 
\begin{equation}
   \loss{total} = \loss{tex} + \loss{mask} + \loss{edge} + \loss{lap}
   \label{eq:total}
\end{equation}
The texture reconstructions loss $\loss{tex}$ is defined as the sum of an $\loss{1}$ loss and perceptual distance metric $\loss{perc}$~\cite{zhang2018perceptual}:
\begin{equation}
    \loss{tex} = \sum_i |I^r_i-I_i| + \loss{perc}(I^r_i,I_i)
    \label{eq:tex}
\end{equation}
The mask reconstruction loss combines an MSE loss and a bi-directional distance transform loss (see details in Appendix).
\begin{equation}
    \loss{mask} = \sum_i ||A^r_i-A_i||^2_2 + \loss{bi-dt}(A^r_i,A_i)
    \label{eq:mask}
\end{equation} 
In addition to reprojection losses in Eq.~\ref{eq:tex} \&~\ref{eq:mask}, we employ smoothness reguralizers on the mesh: $\loss{edge} = ||E - l||^2_2$ is an MSE loss penalizing edge lengths that deviate from the mean initial edge length $l$, while $\loss{lap} = ||L_{\textrm{cot}}V||_2$ is a cotangent-laplacian loss that minimizes mean curvature~\cite{nealen2006laplacian}.

\Paragraph{Initialization and Warmup} 
We initialize cameras with the noisy input cameras, $V_0$ with an ico-sphere and $\Delta V$ with zeros. 
During an initial warmup phase of 500 iterations, we freeze cameras and optimize shape without the texture loss. 
We start with a very low-resolution sphere and subdivide it twice during warmup, at 100 and 300 iterations respectively. 

\Paragraph{Texture Sampling} 
We compute the texture map $T$ after each shape update during optimization. 
For each training view $i$, and for each pixel $p$, we find the $K$ closest faces intersecting a ray originating at $p$ and the corresponding points of intersection $x_{1..K}$.
We compute texels $c_k = T(x_k)$, described in Sec.~\ref{sec:tex-transfer}, and set $w_i=0$ in Eq.~\ref{eq:textrans} so that image $I_i$ does not contribute to the texture for pixel $p$ in the rendered $i$-th view. 
This ensures that image $I^r_i$ for camera $\pi_i$ is generated by sampling colors from all images $I_{1..N}$ but $I_i$ to encourage photometric consistency. 

\Paragraph{Handling Variable Topology} 
Each gradient descent step updates the vertex positions of the mesh and the camera parameters.
However, the topology of the shape is left intact. 
To handle objects with varying topology and to deviate from shapes homeomorphic to spheres, we update the topology of our shapes during optimization.
At intermediate steps during training, we voxelize our mesh~\cite{binvox, nooruddin03simplification}, project voxels onto the view plans and check for occupancy by comparing to the ground truth silhouettes $A_{1...N}$.
We remove voxels that project to an unoccupied area in any mask.
We re-mesh the remaining voxels using marching cubes, reset all shape-optimization parameters and resume optimization.

\section{Experiments}
\label{sec:exp}

We test our differentiable stereopsis approach, which we call \approachName, on three datasets: Google's Scanned Objects~\cite{google2020scanned}, Tanks and Temples~\cite{knapitsch2017tankstemples} and the Amazon-Berkeley Objects~\cite{collins2022abo}. \revised{We additionally evaluate on DTU MVS~\cite{jensen2014largeDTU} in the Appendix.}
We run extensive quantitative analysis on objects of varying topology and shapes, for which 3D ground truth is available.
We also show qualitative results on real objects and challenging real-world scenes.

\subsection{Experiments on Google's Scanned Objects}

Google's Scanned Objects (CC-BY 4.0)~\cite{google2020scanned} consists of 1032 common household objects that have been 3D scanned to produce high-resolution Lambertian textured 3D meshes. 
From these, we pick 50 object instances with varying shape, topology and texture for quantitative analysis including toys, electronics, instruments, appliances, cutlery and many more.
For each object, we render $2048\times2048$ RGBA images from 12 random camera viewpoints.
Camera rotation Euler angles and field-of-view are uniformly sampled in $[0\degree,360\degree]$ and $[20\degree,50\degree]$ respectively. 
To the cameras, we add rotation noise $\theta \sim \mathcal{N}(0,\sigma^2)$ about a uniformly sampled axis with varying $\sigma=\{10\degree,20\degree,30\degree\}$.

\begin{figure*}[t!]
\begin{center}
   \includegraphics[width=0.33\linewidth]{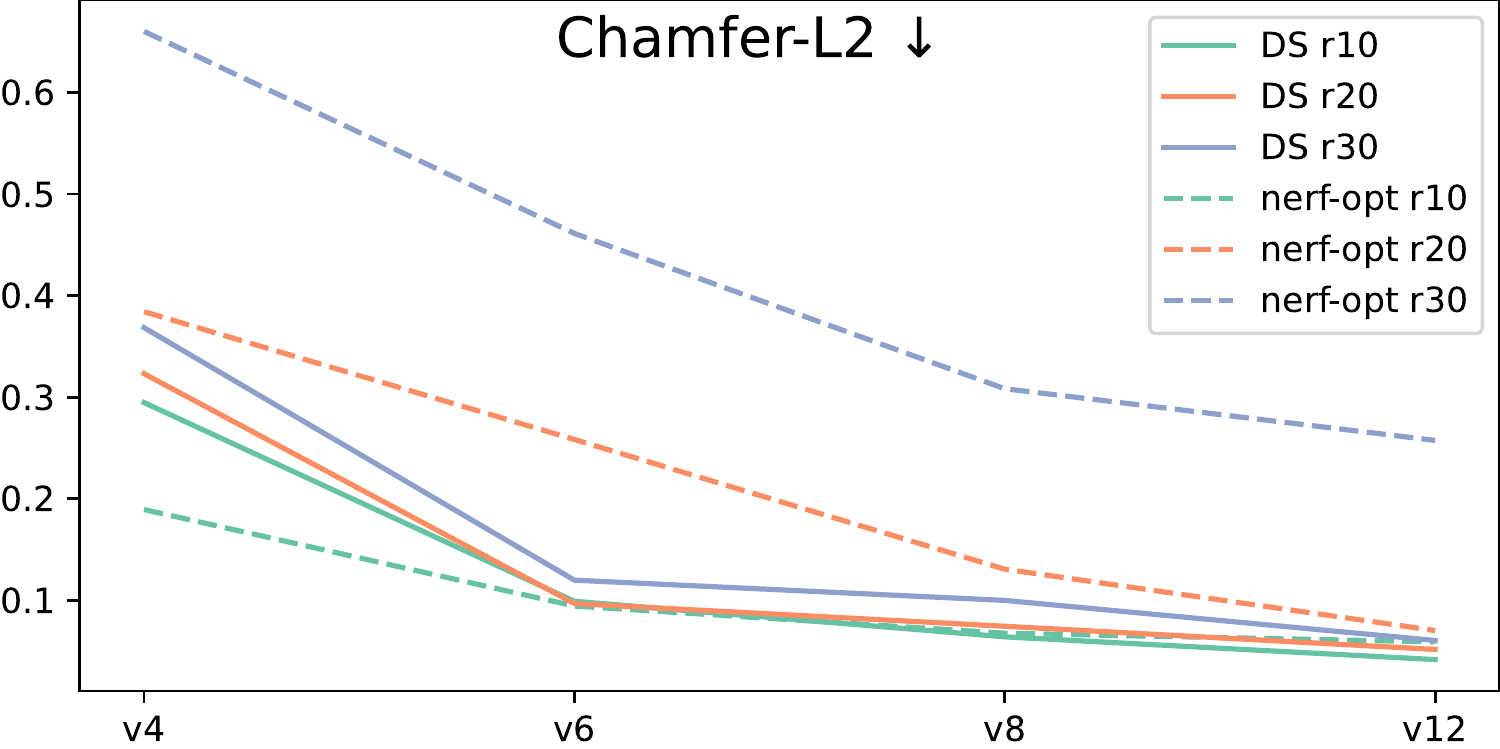}
   \includegraphics[width=0.33\linewidth]{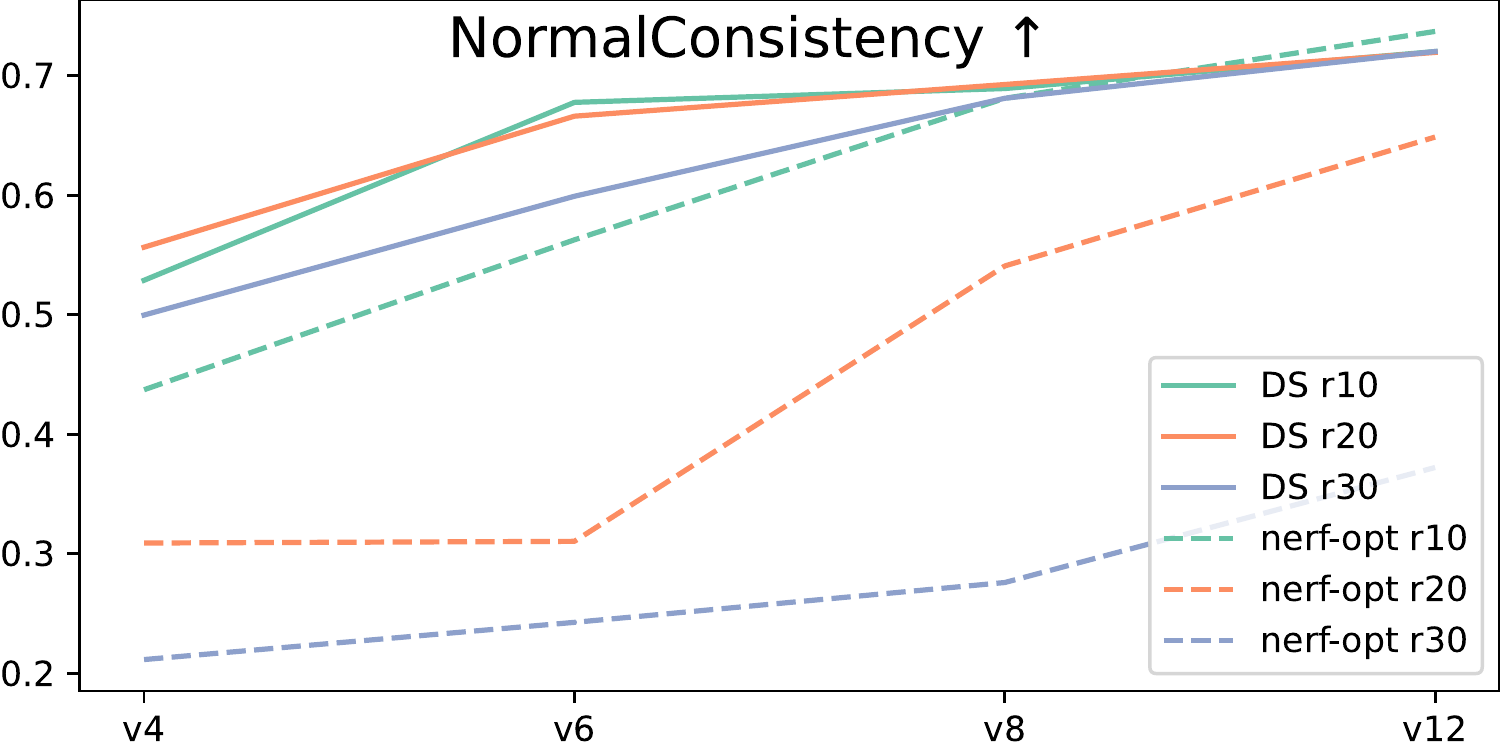}
   \includegraphics[width=0.33\linewidth]{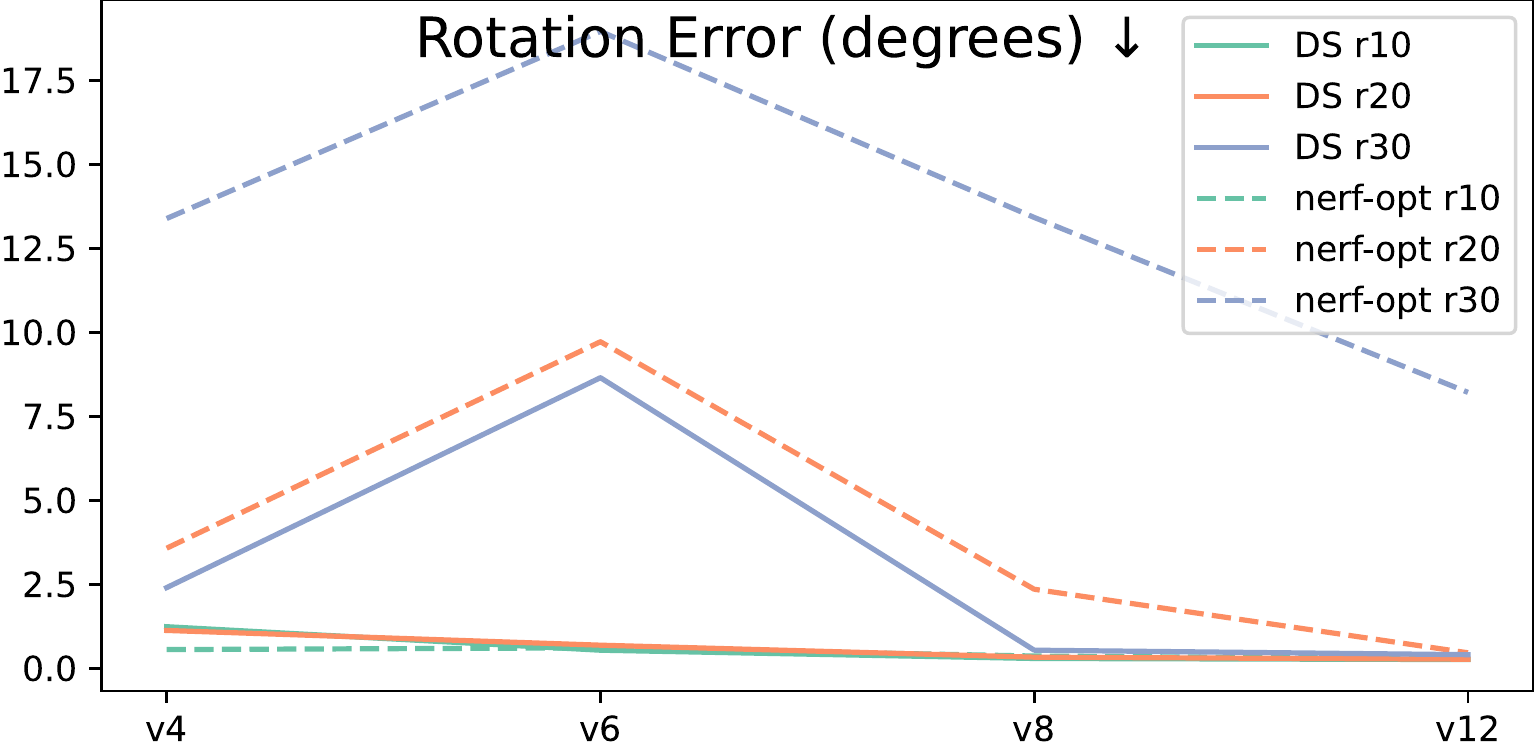}
   \includegraphics[width=0.33\linewidth]{"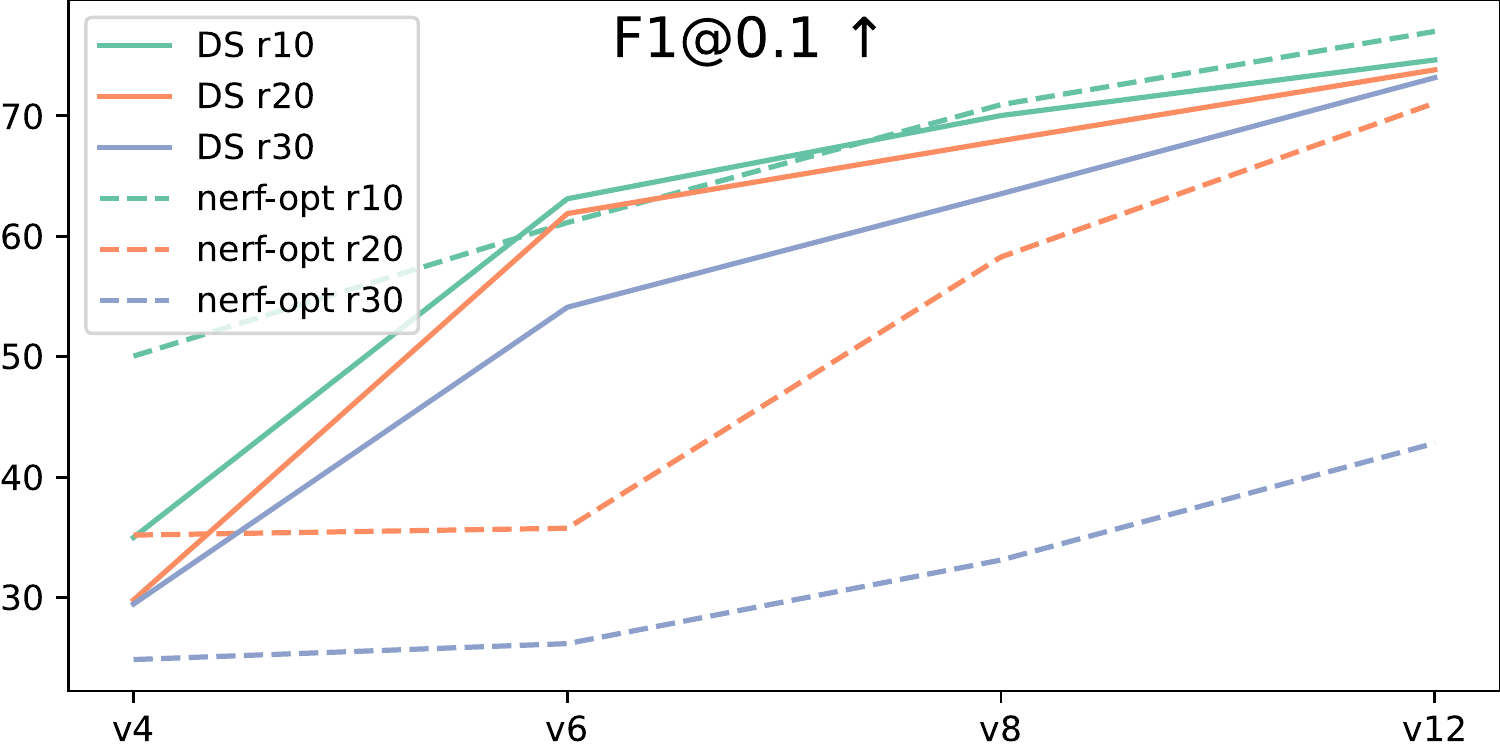"}
   \includegraphics[width=0.33\linewidth]{"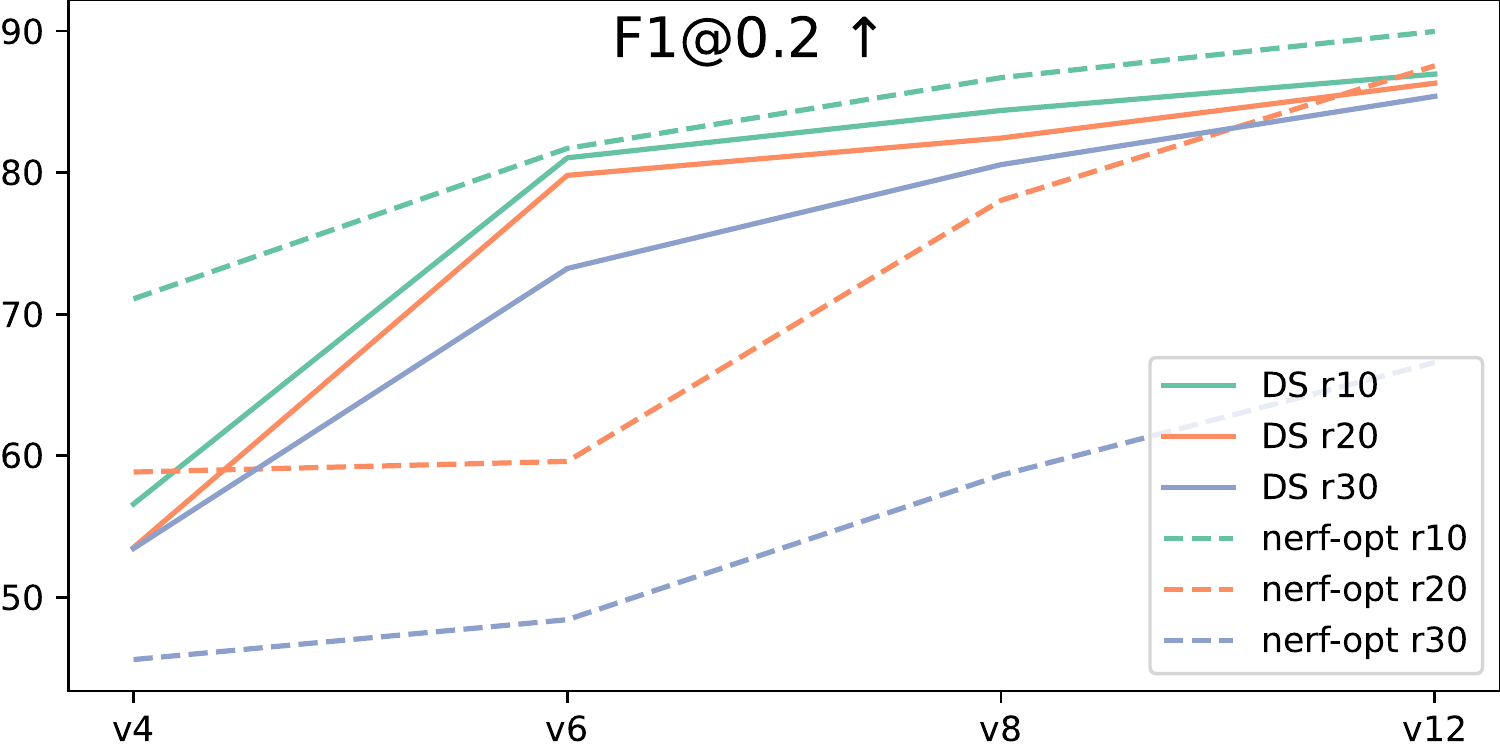"}
   \includegraphics[width=0.33\linewidth]{"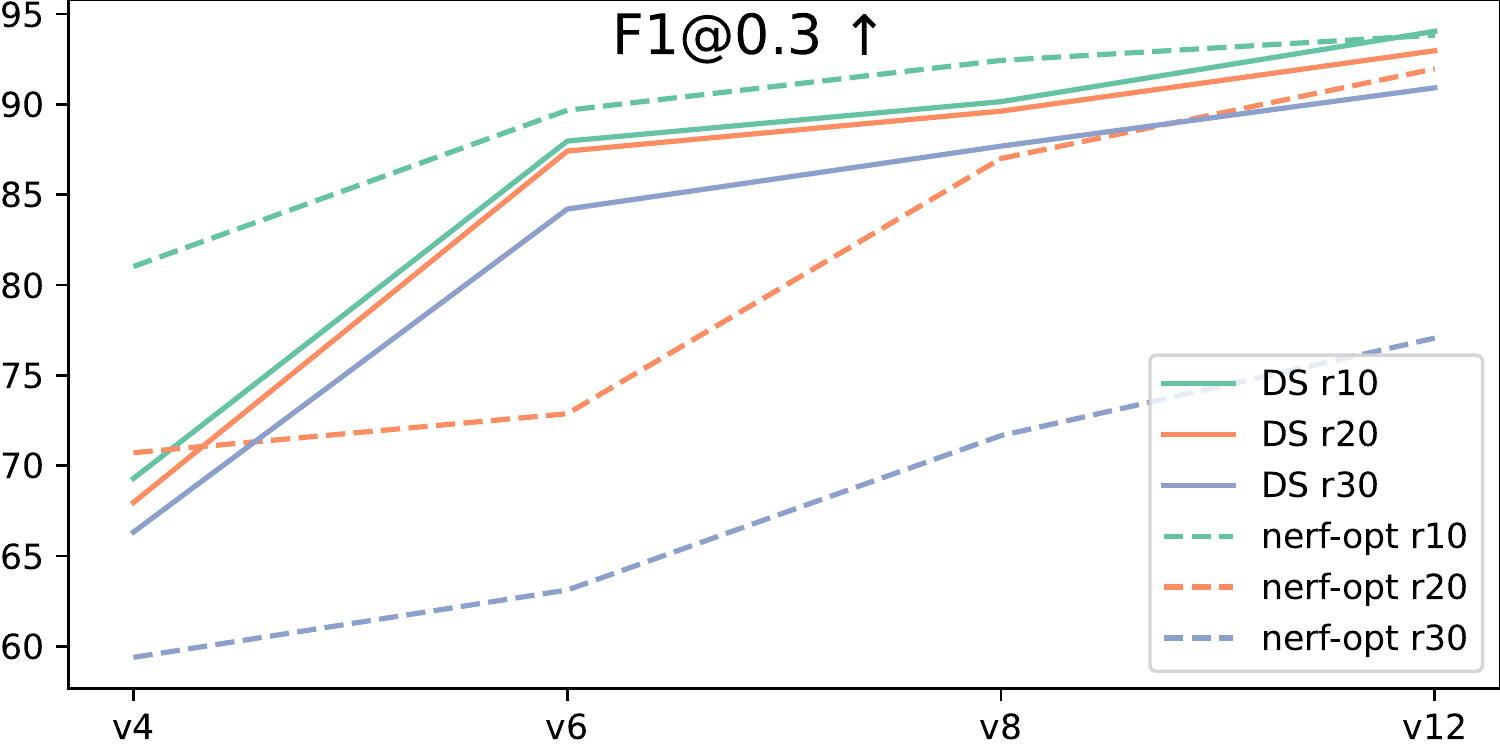"}
\end{center}
    \vspace{-5mm} %
   \caption{Performance of \approachName~and nerf-opt on Google's Scanned Objects with varying number of views $\{4, 6, 8, 12\}$ (x-axis) and camera noise $\{10\degree, 20\degree, 30\degree\}$. 
   Each plot reports the median across the 50 objects in our evaluation set. 
   We report shape reconstruction metrics (Chamfer, F1), normal consistency, and camera error.
   For Chamfer and camera error, lower is better. 
   For everything else, higher is better.}
    \vspace{-4mm} %
\label{fig:quant-line}
\end{figure*}

\Paragraph{Metrics} 
We report a variety of metrics to quantitatively compare the predicted with the ground-truth meshes.
We use $\loss{2}$-Chamfer distance, normal consistency and F1 score at different thresholds, following~\cite{gkioxari2019mesh}.
Since predicted shapes don't lie in the same coordinate frame as the ground-truth, we align predictions to ground truth before benchmarking via the iterative-closest-point (ICP) algorithm~\cite{besl1992icp}. \revised{
See Appendix for more details.
}
Lastly, we report the rotation error (in degrees) between the ground truth and the output cameras from \approachName~optimization.

\Paragraph{Comparison with baselines}
We extensively compare to NeRF~\cite{mildenhall2020nerf}, as the state-of-the-art volumetric method, which learns an implicit function from accurately posed input views.
While NeRF doesn't explicitly output shape, we extract geometry from its implicit representation via voxelization and run marching cubes to get a mesh.
We compare to two NeRF variants which use additional mask information: 
(a) \emph{nerf} - the original NeRF approach with an additional MSE loss on rendered masks, and
(b) \emph{nerf-opt} - which is the same as (a) but optimizes camera poses with gradients from the reprojection loss. 
nerf-opt uses the same camera parametrization as our approach. 
To prevent NeRF from collapsing due to large areas of white background in the input views, all NeRF baselines sample 50\% of their points inside the mask in every iteration. 
We also compare qualitatively to IDR~\cite{yariv2020multiview}, a volumetric method with an implicit representation that learns geometry and appearance from sparse wide-baseline images and masks with noisy camera poses.
In the Appendix, we also compare to COLMAP~\cite{schoenberger2016sfm, schoenberger2016mvs} as the state-of-the-art photogrammetry approach.
Finally, we compare to variants of our approach: (a) \emph{\approachName-notex}, which does not use any texture information removing $\loss{tex}$ from Eq.~\ref{eq:total}; and (b)  \emph{\approachName-naive}, which naively optimizes a UV texture image in addition to shape/camera instead of using our texture-transfer. 
For texturing, the texture image is mapped to the mesh surface using a fixed UV map \cite{hughes2014computer} that is automatically computed with Blender \cite{blender}. 
Whenever the mesh topology changes, the texture image is re-initialized and the UV-map recomputed.

Fig.~\ref{fig:quant-line} quantitatively compares \approachName~to nerf-opt, the best performing NeRF variant of the two. 
We train with varying number of views $N=4,6,8,12$ (x-axis) and varying camera noise  $\{10\degree, 20\degree, 30\degree\}$.
Each plot reports the median across the 50 instances selected from the dataset.
For small camera noise ($10\degree$), \approachName~and nerf-opt (\textcolor{r10}{green lines}) achieve comparable Chamfer and F1, except for $N=4$ views where nerf-opt achieves higher F1.
Undoubtedly, predicting shape from 4 views is challenging for all methods, as indicated by the absolute performance and is the only setting where nerf-opt performs better than \approachName.
For larger camera noise ($20\degree$), \approachName~performs better than nerf-opt (\textcolor{r20}{orange lines}) under all metrics for $N\ge 6$ and on par for $N=4$.
For even larger camera noise ($30\degree$), \approachName~leads by a significant margin (\textcolor{r30}{blue lines}) for all $N$ and all metrics.
We note that as the number of views $N$ increases, both methods converge roughly to the same performance for $10\degree$ \& $20\degree$ noise.
For $30\degree$ noise, \approachName~also converges to the above optimum with increasing views $N$.
On the other hand, nerf-opt is unable to recover shape or cameras for $30\degree$ noise and achieves much lower reconstruction quality.
These results prove that \approachName~can learn better shapes and recover cameras even under larger camera noise and fewer views.
When given slightly more views, \approachName~reaches the same reconstruction quality as with little camera noise proving its robustness to errors in cameras. 
\revised{See Appendix for quantitative comparisons to IDR, COLMAP, and DS-naive.}

\begin{figure}[t]
   \centering
   \includegraphics[width=0.99\linewidth]{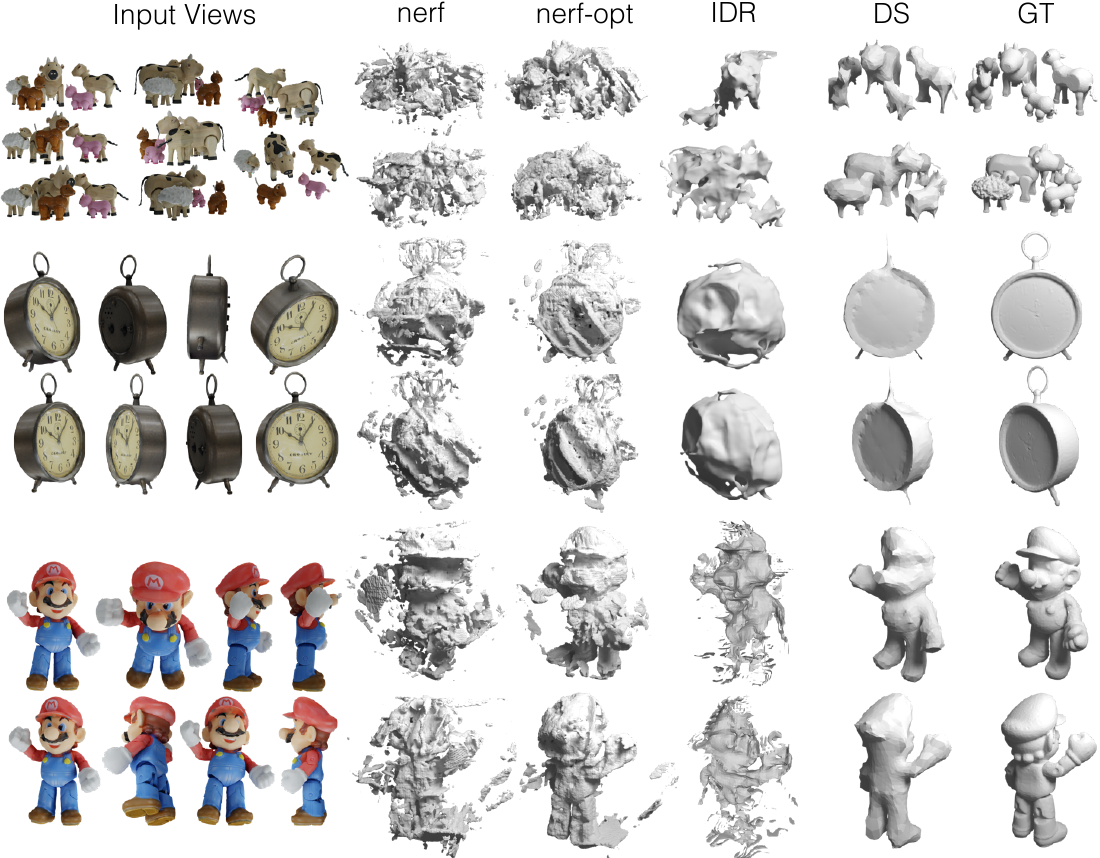}
  \vspace{-3mm} %
   \caption{
   Results of nerf, nerf-opt, IDR and \approachName~with $8$ views and $30\degree$ camera noise. nerf-opt and IDR outperform nerf but they fail to capture good shape. \approachName~captures better geometry illustrating its robustness to high levels of camera noise.
   }
  \vspace{-4mm} %
    \label{fig:qual-nerfopt}
\end{figure}

Fig.~\ref{fig:qual-nerfopt} qualitatively compares nerf, nerf-opt, IDR and \approachName~with $8$ views and $30\degree$ noise.
IDR and both NeRF variants produce shapes with cloudy artifacts, with nerf-opt visibly outperforming nerf.
\approachName~captures better shape under the same settings proving its robustness to noisy cameras and few views. 
We observe that in a few-view wide-baseline setting, like ours, implicit volumetric approaches attempt to explain the few input views without relying on accurate shape geometry and appearance. However, meshes, which explicitly represent surfaces, offer stronger surface regularization and predict more precise geometry.
Also, in the first row of Fig.~\ref{fig:qual-nerfopt} we observe that \approachName~is able to reconstruct different animals as disconnected components despite having been initialized to a single sphere.

\begin{figure}[b]
   \centering
  \vspace{-4mm} %
    \includegraphics[width=0.95\linewidth]{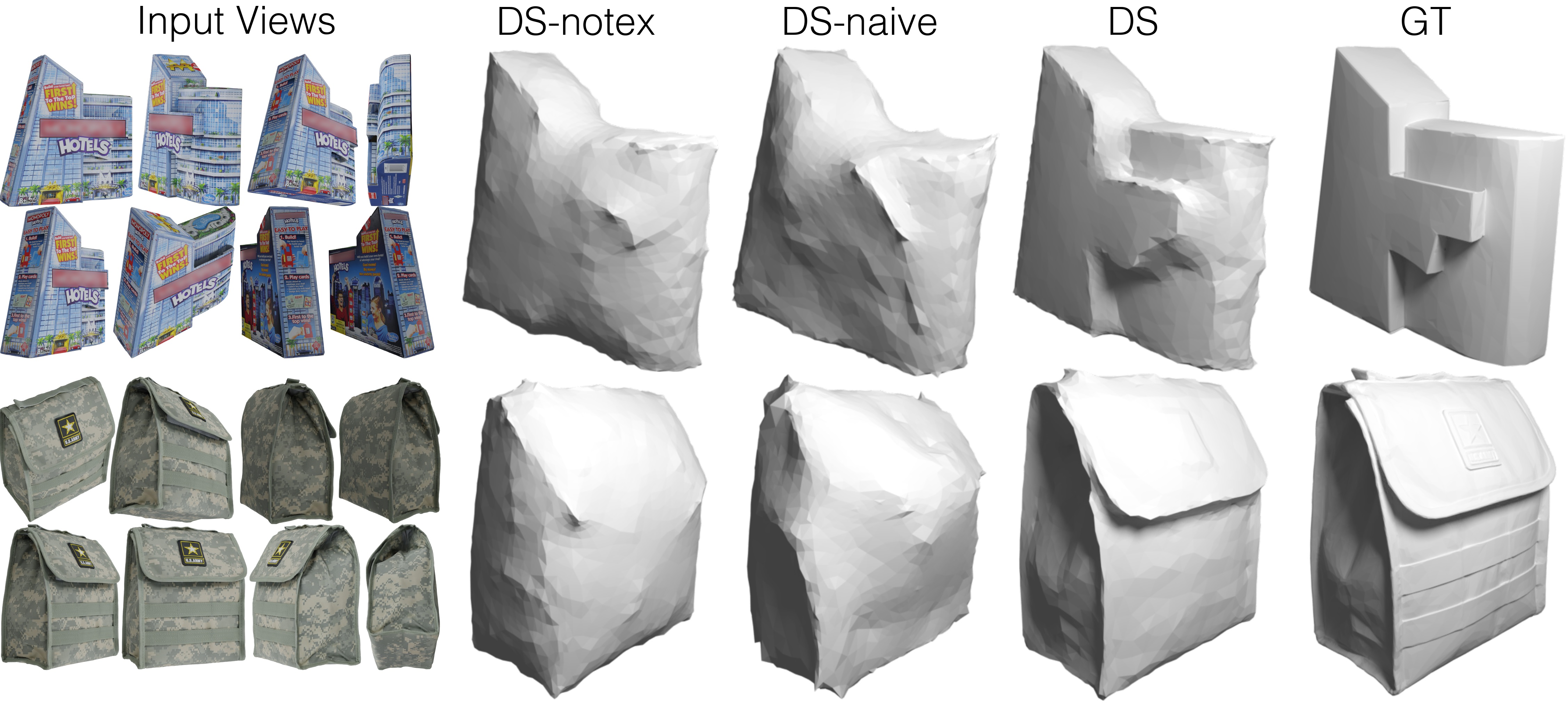}
   \vspace{-3mm} %
   \caption{\approachName~without texture (\approachName-notex), \approachName-naive and \approachName~with $8$ views and $20\degree$ camera noise. \approachName-notex fails to capture shape concavities, while \approachName-naive fails to recover accurate shape and cameras. The ground truth shape (GT) is shown in the last column.
   }
  \vspace{-4mm} %
\label{fig:qual-notex}
\end{figure}

Fig.~\ref{fig:qual-notex} compares \approachName~to \approachName~without texture (\emph{\approachName-notex}) and naive texture map optimization (\emph{\approachName-naive}) with $8$ input views and $20\degree$ camera noise. 
\approachName-notex fails to capture shape concavities, which are impossible to capture via just silhouettes.
\approachName-naive results in shapes with some concavities but in the wrong place and of shape quality similar to \approachName-notex. With naive texture optimization as in \approachName-naive, texture converges to the mean texture from different images, providing unreliable gradients to improve shape/cameras and leading to suboptimal geometry. In contrast, \approachName~accurately captures creases in object shapes by exploiting texture.

Fig.~\ref{fig:qual-google} shows qualitative results on Google's Scanned Objects. 
For each object, we train with $8$ input views and $20\degree$ camera noise.
We show the input views (left) and the output shape and texture for two novel views (right).

Fig.~\ref{fig:shapechange} shows the evolution of shape with time for two examples from Google's Scanned Objects with $8$ input views and $20\degree$ camera noise.
We remesh the shape by updating its topology at three intermediate steps during optimization.
This brings the final shape close to the ground truth both in terms of geometry and topology.

We also show failure modes in the Appendix.

\subsection{Results on Amazon Products}
We show results from images of 6 real-world objects from the ABO dataset (CC-BY-NC 4.0)~\cite{collins2022abo}.
Pixel-thresholding the white-background images gives masks. 
Camera poses for these images are unknown and COLMAP fails to give sensible estimates.
We get rough initial cameras by manually annotating a set of 40 keypoint correspondences across all images of an object.
We estimate the parameters for a weak-perspective camera for each image using a orthographic rigid-body factorization formulation~\cite{marques2009estimating} adopted in \cite{kanazawa2018learning, kar2015category, vicente2014reconstructing}. 
We initialize our perspective cameras using the computed weak-perspective cameras and assume a $30\degree$ field-of-view.

\begin{figure}[b]
   \centering
  \vspace{-3mm} %
   \includegraphics[width=0.99\linewidth]{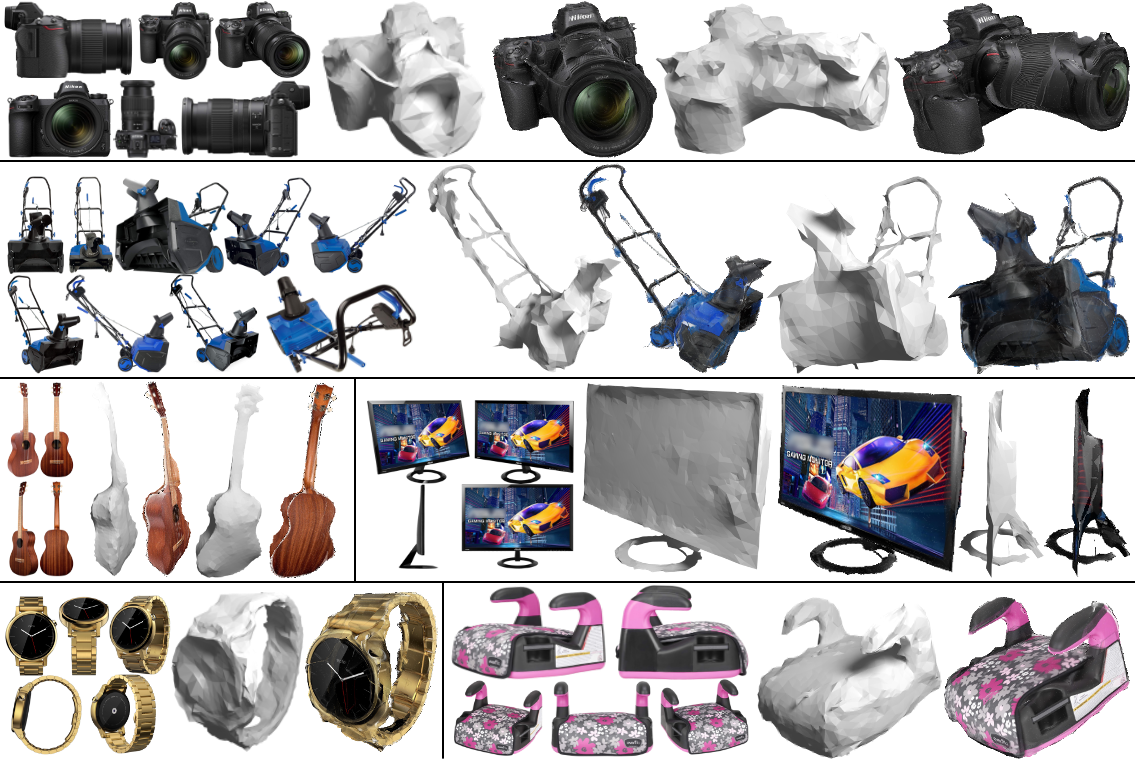}
  \vspace{-3mm} %
   \caption{\approachName~evaluated on real-world product images from Amazon~\cite{collins2022abo}. For each example, we show input views (left) and reconstructed shape and texture for novel views (right).}
  \vspace{-3mm} %
\label{fig:qual-amazon}
\end{figure}

\begin{figure*}
\begin{center}
  \includegraphics[width=0.99\linewidth]{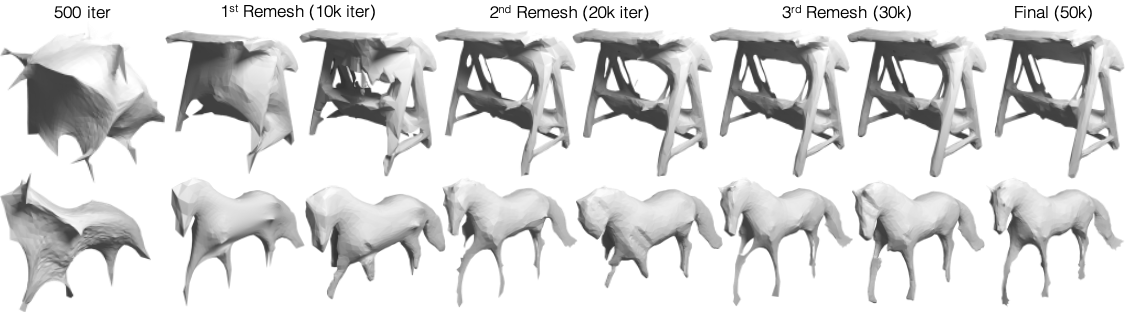}
\end{center}
  \vspace{-5mm}  %
  \caption{Evolution of shape over time for \textit{Garden Swing} (top) and \textit{Breyer Horse} (bottom) from Google's Scanned Objects with $8$ views and $20\degree$ camera noise. We visualize shape at key optimization steps: at the end of warmup (at 500 iterations), before and after the $\nth{1} / \nth{2} / \nth{3}$ remesh (at 10k/20k/30k iterations) and final shape (at 50k iterations).}
\label{fig:shapechange}
\end{figure*}

Fig.~\ref{fig:qual-amazon} shows shape and texture reconstructions.
Despite very noisy cameras and few views, ranging from $4$ to $9$, our approach reconstructs shape and texture reasonably well even for challenging shape topologies like the lawnmower.
We also note that~\approachName~is able to reconstruct more specular surfaces like the wristwatch in the last row. 

\begin{figure*}
\begin{center}
  \includegraphics[width=0.99\linewidth]{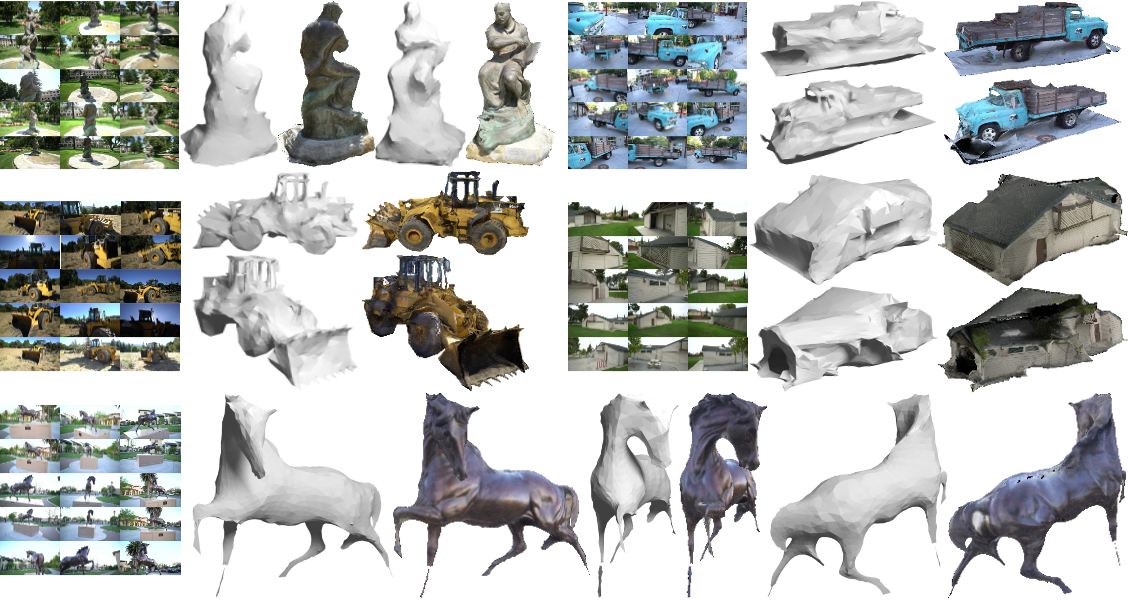}
\end{center}
  \vspace{-5mm}  %
  \caption{Reconstructions of \approachName~on \textit{Ignatius}, \textit{Truck}, \textit{Caterpillar}, \textit{Barn} and \textit{Horse} from Tanks and Temples with $15$ input views and SfM-generated camera poses.
  For each example, we show input views (left), shape and texture reconstructions from two novel views (right).
  Silhouettes for \textit{Horse} were generated by a pretrained off-the-shelf 
  2D object detector.
  }
  \vspace{-4mm}  %
\label{fig:qual-t2}
\end{figure*}

\subsection{Results on Tanks and Temples}
Tanks and Temples (CC-BY-NC-SA 3.0)~\cite{knapitsch2017tankstemples} is a 3D reconstruction benchmark consisting of RGB videos of indoor and outdoor scenes with corresponding laser-scanned ground-truth 3D point clouds.
The dataset comes with cameras computed by COLMAP's SfM pipeline~\cite{schoenberger2016sfm}. 
We evaluate on 7 scenes using \emph{only 15} input images and corresponding SfM-reconstructed cameras as initialization. 
For \textit{Barn}, \textit{Ignatius}, \textit{Caterpillar} and \textit{Truck}, we generate masks by rendering the 3D point clouds from SfM-reconstructed cameras. 
To stress test our approach without relying on 3D point clouds to get masks, for \textit{Horse}, \textit{Family} and \textit{Train} we use an off-the-shelf object detector~\cite{kirillov2019pointrend} pretrained on COCO~\cite{COCO}. 

Fig.~\ref{fig:qual-t2} shows reconstructions for scenes from Tanks and Temples with $15$ input views and SfM-reconstructed cameras.
\approachName~is able to produce good reconstructions and undoubtedly has a harder time for \textit{Barn} due to occlusions by trees.
For \textit{Family} and \textit{Train}, detected masks are poor leading to bad reconstructions. 
\revised{
In the Appendix, we compare to IDR, NeRF-opt, and COLMAP.
}

\section{Discussion}
We propose Differentiable Stereopsis (\approachName) by pairing traditional model-based stereopsis with modern differentiable rendering.
We show results on a diverse set of object shapes with noisy cameras and few input views. 
\revised{
Even though \approachName~performs well, it has limitations. 
It assumes Lambertian surfaces and consistent lighting.
\approachName~works for objects -- extending to complex scenes is future work.
While \approachName~is robust to noisy masks (\eg predictions from Mask R-CNN) and inaccurate cameras provided at input, eliminating them from the input alltogether is important future work.
}

\clearpage

\begin{figure*}
\begin{center}
  \includegraphics[width=0.99\linewidth]{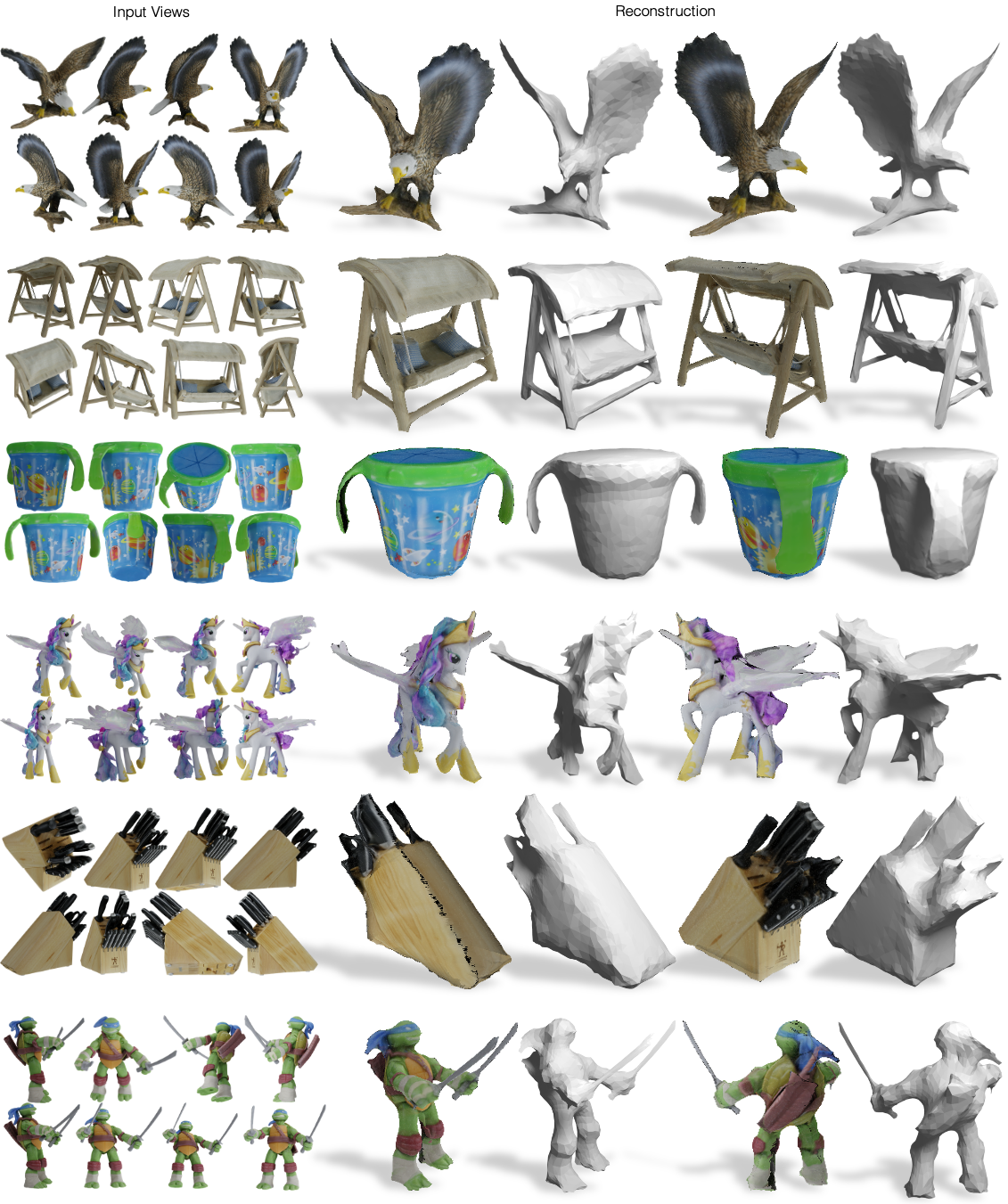}
\end{center}
  \caption{Qualitative results of \approachName~on Google's Scanned Objects with $8$ input views and $20\degree$ camera noise. We show input views (left) and reconstructed shape and texture from two novel views (right).}
\label{fig:qual-google}
\end{figure*}

\clearpage

{\small
\bibliographystyle{ieee_fullname}
\bibliography{main}
}

\clearpage
\appendix

\section{Additional Training Details}
\subsection{Misc}
We downsample the original $2048\times2048$ images to $400\times400$ for all methods but COLMAP. Both NeRF baselines are trained for 100k iterations and take $\sim$ 10 hours to run on a 32GB NVIDIA V100. DS is trained for 50k iterations and takes $\sim$ 9 hours to run on the same GPU.

\subsection{Bi-directional distance transform loss}
This section describes the bi-directional distance transform loss $\loss{bi-dt}$ used in the mask reconstruction loss in Equation~6 of the main text. Our novel loss is a differentiable version of a non-differentiable naive bi-directional distance transform loss. We first define the naive non-differentiable loss and then describe our differentiable adaptation to it. For convenience, we misuse $A$ (or $A^r$) to refer to the set of pixels where GT mask $A$ (or the rendered mask $A^r$) is 1. 

A naive non-differentiable bi-directional distance transform loss $\loss{bi-dt-naive}(A^r, A)$ between the rendered mask $A^r$ and GT mask $A$ consists of 2 components. The first penalizes pixels $p\in \ A^r \setminus A$, where the rendered mask is 1 but GT mask is 0, by the distance of $p$ from $\texttt{NN}(p,A)$ - the closest occupied pixel in the GT mask. $\texttt{NN}$ is the nearest neighbor operation in euclidean space. The second component penalizes pixels $q \in \ A \setminus A^r$, where the GT mask is 1 but rendered mask is 0, by the distance of $q$ from $\texttt{NN}(q,A^r)$ - the closest occupied pixel in the rendered mask. 

\begin{equation}
\loss{bi-dt-naive}(A^r, A) = \sum_{p\in A^r\setminus A} \texttt{NN}_d(p,A) +  \sum_{q\in A \setminus A^r} \texttt{NN}_d(q,A^r))
\end{equation}
where $\texttt{NN}_d(p,A) = d(p,\texttt{NN}(p,A))$ is the distance between $p$ and its nearest neighbor in $A$ as measured by a metric $d$.

The operation of finding the set of pixel locations where the rendered mask $A^r$ is occupied is the only non-differentiable one in the loss above. Our novel adaptation provides a differentiable approximation for the same. Given a pixel $p$ in $i$-th image $I_i$, recall (from Section~3.1) that the rasterizer finds $K$ points $x_{1..K}$ on the surface of the mesh that project to $p$ under camera $\pi_i$. Therefore, $\pi_i(x_k)$ is a differentiable approximation to $p$ for each $x_k$. We approximate $p \sim \hat{p} = \sum_k w_k\pi_i(x_k)$ as a normalized weighted sum of projections $\pi_i(x_k)$ with weights from the softmax blending that composites texels $c_{1..K}$ into a final color at $p$. Let $\hat{A^r} = \{\hat{p} | p \in A^r; \hat{p}=\sum_k w_k\pi_i(x_k) \}$. The full differentiable bi-directional distance transform loss is

\begin{equation}
\loss{bi-dt}(A^r, A) = \sum_{p\in A^r\setminus A} \texttt{NN}_d(\hat{p},A) +  \sum_{q\in A \setminus A^r} \texttt{NN}_d(q,\hat{A^r}))
\end{equation}

where $d(x,y) = \texttt{clamp}(||x-y||_2, \tau_\text{min}, \tau_\text{max})$ is the clamped L-2 euclidean distance. We set $\tau_\text{min}$ to 2 pixels and $\tau_\text{max}$ to $0.1$ of the shorter image dimension.

\subsection{Learning Schedule}
We run our optimization for $50$k iterations using SGD with a momentum of $0.9$ and an initial learning rate of $0.01$. We use cosine annealing for our learning rate with warm restarts~\cite{loshchilov2016sgdr} which we find helps avoid local minima for both shape and cameras. We clip gradient norms for stability. For mesh rasterization, we use PyTorch3D \cite{ravi2020pytorch3d} with $K=6$ and a blur radius that decays exponentially from $5\times10^{-5}$ to $10^{-6}$ over the course of optimization.

\subsection{Aligning meshes for benchmarking}
\revised{
As cameras are optimized, ground-truth and optimized shapes are not in a common coordinate space and must be aligned before benchmarking. However, different benchmarking metrics are minimized by different alignments. Therefore, 
for every instance, for each metric, we find 3 possible alignments and pick the best. The first is a brute force minimization of Chamfer-L2 over scale/depth in camera~$0$'s view coordinate space. The second and third additionally optimize rotation and translation via ICP from ground-truth to predicted mesh and vice-versa.
}

\section{Additional Results}
\subsection{COLMAP}
\label{sec:colmap}
In Fig.~\ref{fig:qual-colmap-t2}, we show dense pointcloud reconstructions by COLMAP on scenes from the Tanks and Temples dataset as the number of views reduces from 100 to 50 to 25 to 15. The pointcloud density drops drastically as we reduce the number of views below 50. With 15 views, the reconstruction is practically empty and the scene is unrecognizable. 
Our attempts to mesh these points using COLMAP’s Poisson and Delaunay meshers failed. In contrast, as seen in 
Fig.~8 of the main text, 
our approach is far more robust with the same views.

In Fig.~\ref{fig:qual-colmap-google}, we show dense pointcloud reconstructions by COLMAP on instances from the Google dataset with 8 input views and ground-truth cameras. COLMAP fails on 2/50 instances and of the 48 instances it works on, it only reconstructs parts of the scene that are textured and visible in close (narrow-baseline) views. It still misses surfaces that are visible in only 2-3 wide baseline views. For example in Fig.~\ref{fig:qual-colmap-google}, COLMAP fails to reconstruct the back of the green backpack which is visible only in 2/8 input views.

\paragraph{Cameras from SfM}
We tried using COLMAP's Structure-from-Motion (SfM) pipeline to ameliorate the need for noisy camera inputs. It often, if not always, fails to register all cameras into a single coordinate space correctly and usually ends up with multiple groups -- each with 2-3 cameras with reasonably accurate relative orientation. The inability to merge the multiple groups into a single coordinate frame made it infeasible to use the COLMAP cameras as initializations. Note that SfM’s failure is not surprising since we are working with 8 white-background views covering 360-degrees of the object. On average, any surface is visible from~$\sim3$ views only. Furthermore, white background images are harder to calibrate than in-context images with background because of the lack of visual overlap in the background.

\revised{
\subsection{Further comparisons}
}
In addition to comparisons with NeRF~\cite{mildenhall2020nerf} and NeRF-opt in the main paper, we compare to COLMAP~\cite{schoenberger2016sfm}, IDR~\cite{yariv2020multiview}, and DS-naive. 
\cref{tab:rebuttal-quant-r30v8} shows results on GSO following the evaluation protocol of the main paper. 
All methods use 8 views and $\sigma=30\degree$ camera noise, except COLMAP, which uses ground-truth cameras. 
For COLMAP, we compute metrics on the reconstructed dense point cloud via uniform random sampling.

DS outperforms all baselines across all metrics. 
IDR gets slightly better shapes but worse cameras than NeRF-opt. 
However, neither can recover accurate geometry under camera noise.
Surprisingly, DS-naive performs similar to NeRF-opt but with better Chamfer and Normal Consistency, suggesting the mesh representation is robust in noisy settings. 
We note that COLMAP has extremely high precision (99.8\%, compared to DS's 88.4\%) but very poor recall (35.5\%, compared to DS's 78.2\%) at a threshold of 0.2.
This suggests that COLMAP reconstructs accurate point clouds but misses large parts of the shape, supporting the qualitative claim in \cref{sec:colmap}.
The comparison with COLMAP, a method that finds view correspondences followed by triangulation, suggests that such methods have a hard time reconstructing shapes from limited views.

\begin{figure}
  \centering
  \includegraphics[width=0.99\linewidth]{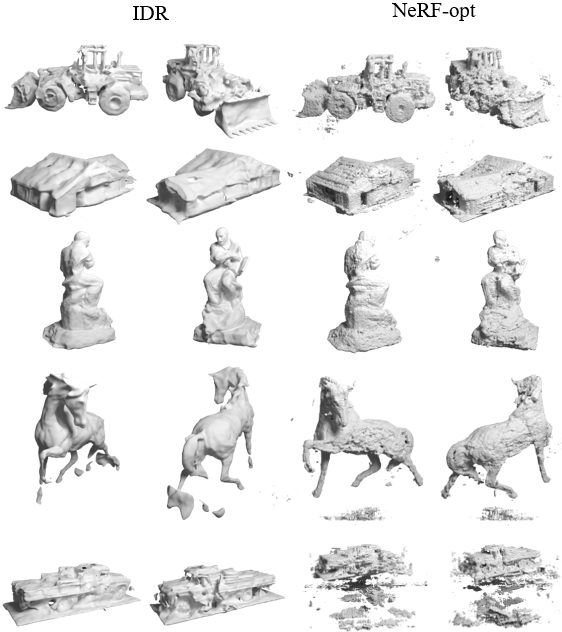}
   \caption{IDR and NeRF-opt on Tanks and Temples with 15 views.}
   \label{fig:rebuttal-qual-T2}
\end{figure}

\paragraph{Tanks and Temples} \cref{fig:rebuttal-qual-T2} compares on Tanks and Temples with 15 views and no camera noise (to be compared to \cref{fig:qual-t2} in the main paper). \cref{tab:fig:quant-t2} shows the corresponding quantitative comparison following the evaluation protocol of the main paper. However, since the ground-truth pointclouds are hollow (without the bottom), the reported numbers only approximate the quality of the shape.
NeRF-opt produces lower quality shapes than DS.
With more views and no camera noise, IDR performs slightly better than DS.
This is not a surprise.
Our approach has an advantage when few views ($\le 12$) are available and cameras are noisy.
But our approach, even with more views ($=15$) and no camera noise, reconstructs shapes well, performing better than NeRF-opt and similar to IDR.

\vspace{-2pt}

\begin{table}
  \centering
  \begin{tabular}{@{}clccccc@{}}
    \toprule
    Scene                        & Method   & Chamfer$\downarrow$ & F1@.1$\uparrow$ & F1@.2$\uparrow$  \\
    \midrule
    \multirow{3}{*}{Barn}        & IDR      &   \textbf{0.14}  &  \textbf{47.6}  &   \textbf{70.2}     \\          %
                                 & NeRF-opt &   0.98  &  33.4  &   48.2     \\          %
                                 & DS       &   0.39  &  35.1  &.  56.0     \\ \hline   %
    \multirow{3}{*}{Caterpillar} & IDR      &   \textbf{0.06}  &  \textbf{59.6}  &   \textbf{81.9}     \\
                                 & NeRF-opt &   0.15  &  54.5  &   77.7     \\
                                 & DS       &   0.07  &  55.8  &   76.4     \\ \hline
    \multirow{3}{*}{Ignatius}    & IDR      &   \textbf{0.18}  &  \textbf{68.2}  &   \textbf{86.9}     \\
                                 & NeRF-opt &   0.20  &  54.6  &   74.5     \\
                                 & DS       &   0.30  &  53.4  &   75.8     \\ \hline
    \multirow{3}{*}{Truck}       & IDR      &   \textbf{0.06}  &  \textbf{54.8}  &   \textbf{79.7}     \\
                                 & NeRF-opt &   1.40  &  23.9  &   40.8     \\
                                 & DS       &   0.14  &  52.0  &   70.7     \\
    \bottomrule
  \end{tabular}
  \caption{Results on Tanks and Temples scenes with 15 input views.}
  \label{tab:fig:quant-t2}
\end{table}

\begin{table}
  \centering
  \begin{tabular}{@{}lccccc@{}}
    \toprule
    Method      & Chamfer$\downarrow$   & F1@.1$\uparrow$ & F1@.2$\uparrow$ & NC$\uparrow$    & Rot$\downarrow$   \\
    \midrule
    COLMAP      & 0.38      & 35.2  & 52.3  & -     & -     \\
    IDR         & 0.52      & 22.0  & 40.5  & 0.23  & 22.8  \\ %
    NeRF-opt    & 0.31      & 33.1  & 58.6  & 0.28  & 13.4  \\ %
    DS-naive    & 0.22      & 29.5  & 53.0  & 0.54  & 14.7  \\ %
    DS          & \textbf{0.10} & \textbf{63.5} & \textbf{80.6} & \textbf{0.68} & \textbf{0.54} \\ %
    \bottomrule
  \end{tabular}
  \caption{Results on GSO with 8 views and $\sigma=30\degree$ camera noise. NC=Normal Consistency; Rot=Rotation Error (in degrees).}
  \label{tab:rebuttal-quant-r30v8}
\end{table}

\clearpage

\begin{figure*}[b]
\begin{center}
  \includegraphics[width=0.99\linewidth]{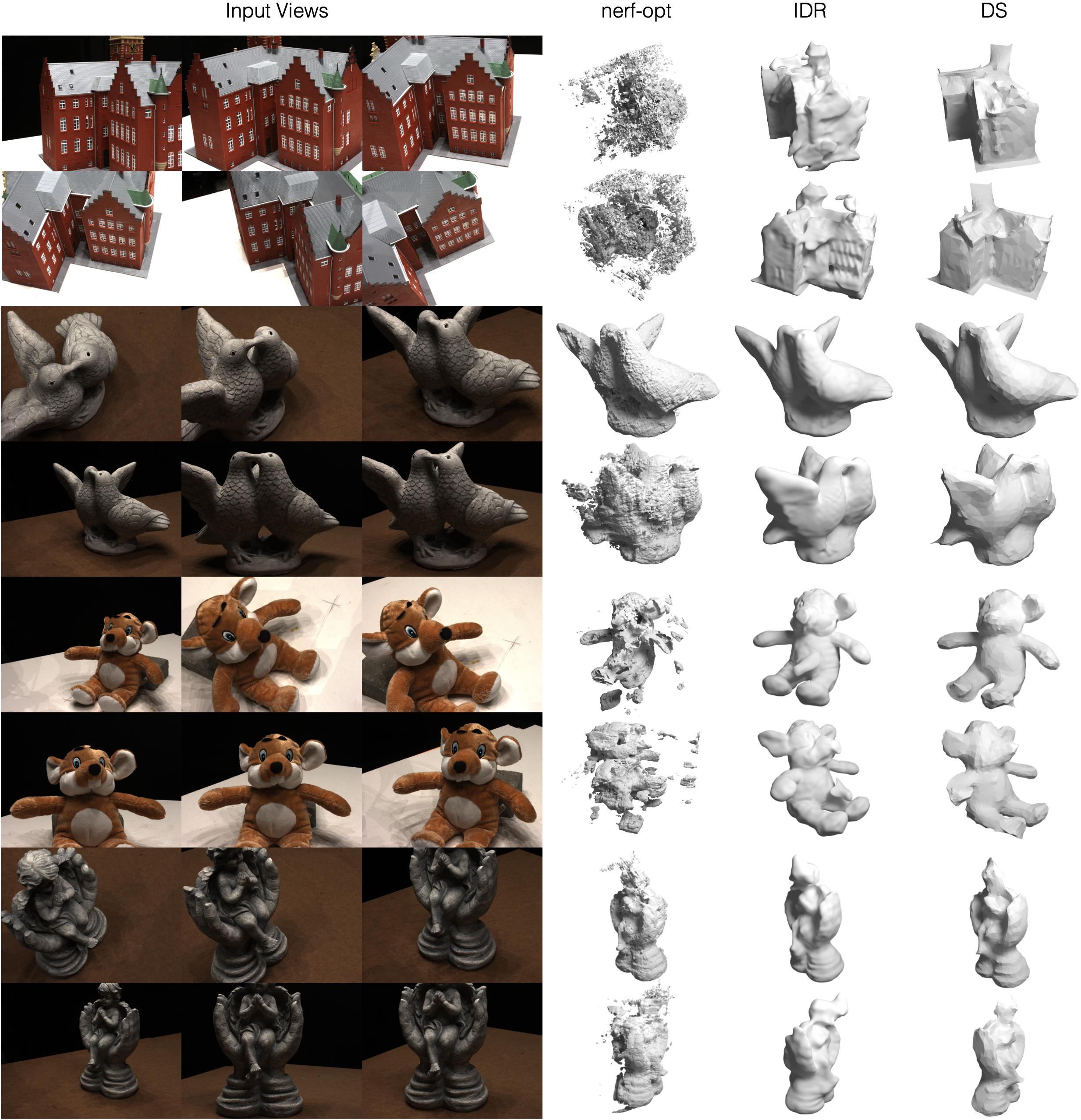}
\end{center}
  \caption{Results on DTU with 6 views and linear camera noise. Figure compares DS to IDR and NeRF-opt.}
\label{fig:qual-dtu}
\end{figure*}

\paragraph{DTU dataset}
\cref{fig:qual-dtu} compares on scenes from the DTU dataset with 6 views and linear camera noise following IDR \cite{yariv2020multiview}. Note that this data is easier to reconstruct because the 6 views span only $~1/\nth{8}$ of the entire viewing sphere and the cameras have an average camera noise of only $\sim1\degree$, which is much lesser than most of our experiments on GSO. We observe that both DS and IDR outperform NeRF-opt, which has cloudy artifacts. IDR works well in this setting with few views and almost-correct cameras. This is not a surprise as DS has an advantage with few views and noisy cameras. IDR reconstructions are slightly more detailed than DS at some places (\eg at the windows of the house and the feet of the bird). However, IDR shapes are less constrained and contain erroneous blobs (\eg at the belly of the teddy bear, the middle and side of the house, and the head of the boy) because of the lack of a sufficient number of views.

\subsection{DS}
\begin{figure*}[h!]
\begin{center}
  \includegraphics[width=0.99\linewidth]{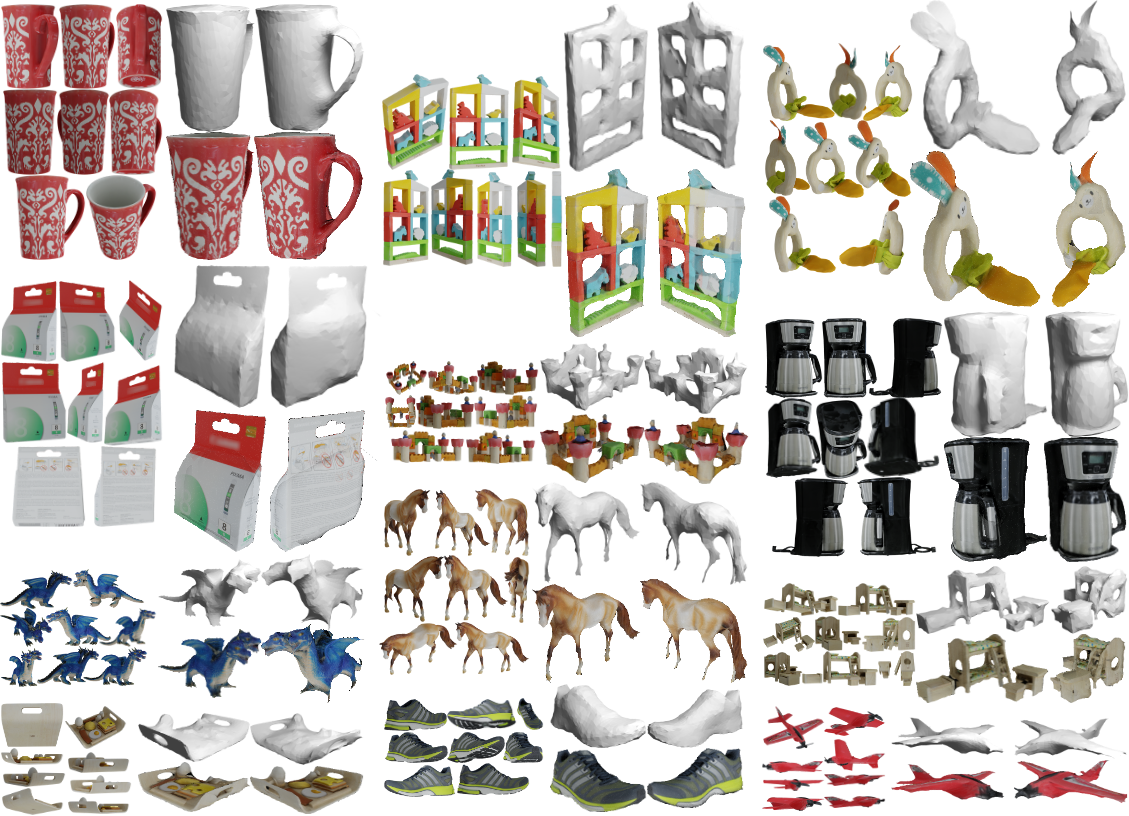}
\end{center}
  \caption{Qualitative shape and texture reconstructions by DS using 8 views and $20\degree$ camera noise for more instances from Google's Scanned Objects Dataset}
\label{fig:qual-google-more1}
\end{figure*}

In \cref{fig:qual-google-more1} 
, we show shape and texture reconstruction from DS on more instances in the GSO Dataset.

\begin{figure*}[h!]
\begin{center}
  \includegraphics[width=0.99\linewidth]{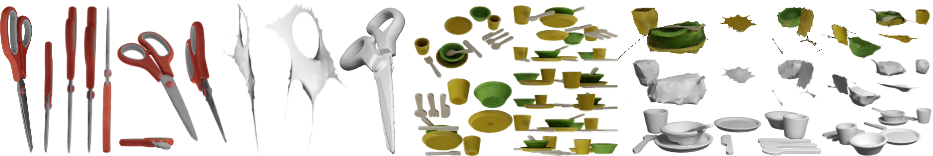}
\end{center}
  \caption{Failure modes for two instances from Google's Scanned Objects. For the leftmost example, we show the input views (left) and reconstructions (middle) and ground truth shape (right). For the rightmost example, we show input views (left), reconstructions (right top) and ground truth shape (right bottom).}
\label{fig:qual-failure}
\end{figure*}

\begin{figure*}
\begin{center}
  \includegraphics[width=0.98\linewidth]{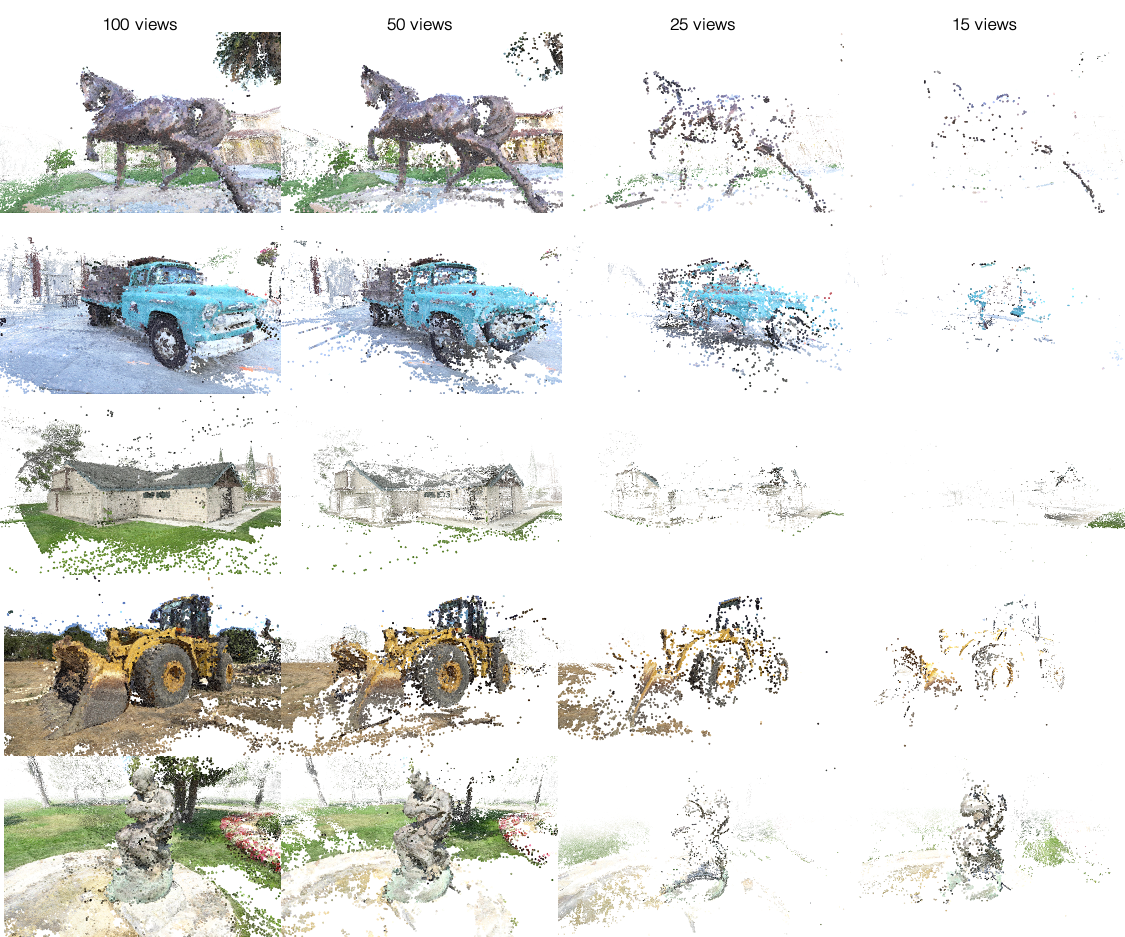}
\end{center}
  \caption{COLMAP-reconstructed dense pointclouds with varying number of input views from scenes in the Tanks and Temples dataset.}
\label{fig:qual-colmap-t2}
\end{figure*}

\begin{figure*}
\begin{center}
  \includegraphics[width=0.99\linewidth]{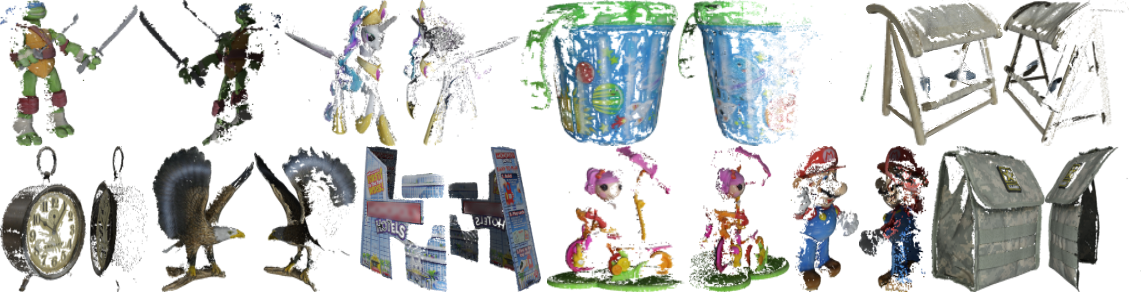}
\end{center}
  \caption{COLMAP-reconstructed dense pointclouds with 8 input views and ground-truth cameras for objects in the Google Scanned Objects Dataset.}
\label{fig:qual-colmap-google}
\end{figure*}

In \cref{fig:qual-failure}, we show some failure modes of DS. Thin structures and small objects are particularly hard to reconstruct. This is because we use a mesh representation with surface smoothening regularization. DS does not model surface specularities and lighting and subsequently also struggles with specular surfaces.

\clearpage

\end{document}